\documentclass{article} %
\usepackage{iclr2027_conference}
\iclrfinalcopy
\usepackage{iftex}
\ifPDFTeX
  \usepackage{times}
  \usepackage[T1]{fontenc}
\else
  \usepackage{fontspec}
\fi

\usepackage{amsmath,amsfonts,bm}

\def\eqref#1{equation~\ref{#1}}

\def\1{\bm{1}}

\DeclareMathAlphabet{\mathsfit}{\encodingdefault}{\sfdefault}{m}{sl}
\SetMathAlphabet{\mathsfit}{bold}{\encodingdefault}{\sfdefault}{bx}{n}

\usepackage{hyperref}
\usepackage{url}
\usepackage{booktabs}     %
\usepackage{longtable}    %
\usepackage{amsmath,amssymb}
\usepackage{graphicx}
\usepackage{multirow}
\usepackage{xcolor}
\usepackage{subcaption}
\usepackage{placeins} 

\newcommand{\ours}{\textsc{RegexRoute}}

\title{Routing Without Embeddings: Fast and Interpretable Routing with Regular Expressions}

\author{%
  \textbf{Yifan Lu\thanks{Equal contribution.}, Qiyue Zhang\textsuperscript{\ensuremath{*}}, Haotian Shan, Hanjie Chen, Jiarong Xing} \\
  Rice University \\
  \texttt{\{yifan.lu, jxing\}@rice.edu}
}

\date{}
\hypersetup{
  hidelinks,
  pdftitle={Routing Without Embeddings: Fast and Interpretable Routing with Regular Expressions},
  pdfauthor={Yifan Lu, Qiyue Zhang, Haotian Shan, Hanjie Chen, Jiarong Xing}
}

\begin{document}
\maketitle
\fancyhead{}
\renewcommand{\headrulewidth}{0pt}

\begin{abstract}
Large Language Model (LLM) routers commonly rely on neural query embeddings, with larger encoders expected to better capture query intent and difficulty.
Yet scaling Qwen2.5 encoders from 0.5B to 72B parameters brings little improvement in routing accuracy (Figure~\ref{fig:encoder-scaling}b)
, suggesting that small encoders may already capture the query properties needed for routing.
We therefore investigate which properties matter and whether they can be extracted directly from text without a neural encoder.
We introduce \ours{}, a pipeline that uses sparse autoencoders (SAEs) to discover interpretable regular-expression (regex) features.
Using unlabeled text, an LLM turns descriptions of grouped SAE latents into regex extractors and refines them to match latent activation patterns.
These extractors supply numerical features to a lightweight routing head, eliminating neural encoding at inference (Figure~\ref{fig:encoder-scaling}a).
Across four benchmarks, one fixed set of 128 features achieves 76.43\% average routing accuracy, comparable to 76.41\% for the strongest neural text encoder baseline, with much smaller latency and strong robustness.
These findings establish explicit, interpretable text features as a practical basis for designing and understanding LLM routers.
\end{abstract}

\begin{figure}[!ht]
\centering
\includegraphics[width=0.9\linewidth]{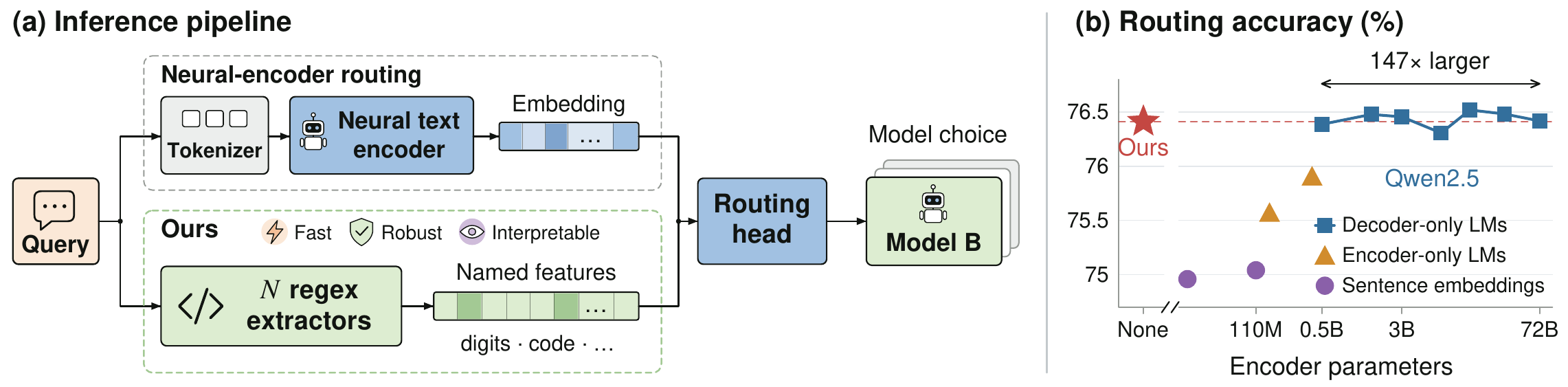}
\caption{(a) Regex extractors replace tokenization and neural encoding at inference. (b) Routing remains competitive, while scaling Qwen2.5 from 0.5B to 72B brings limited gains (Appendix~\ref{app:scaling}). 
}
\label{fig:encoder-scaling}
\end{figure}

\section{Introduction}
\label{sec:intro}

The proliferation of LLM (Large Language Models) with varying capabilities, latency, and price has made \emph{LLM routing} a standard component of LLM service systems \citep{openrouter,notdiamond,martian}.
This need motivated a range of router designs, from nearest-neighbor and clustering methods \citep{zhang2025beyond,jitkrittum2025universalmodelroutingefficient} to learned classifiers \citep{song2025irtroutereffectiveinterpretablemultillm,feng2025graphroutergraphbasedrouterllm,zhuang2024embedllmlearningcompactrepresentations}.
These routers share a common design: a neural text encoder maps each query to a dense \emph{embedding}, and a decision rule in this space selects a model.

Existing routers use neural text encoders with widely varying capacities.
Examples include encoders with roughly 100M \citep{ong2025routellmlearningroutellms,srivatsa2024harnessingpowermultipleminds}, 8B \citep{wang2025iclrouterincontextlearnedmodel,zhang2025beyond}, and even 122B parameters from a deployed language model \citep{varshney2026llmrouterrethinkingrouting}.
The larger encoders are intuitively appealing for routing because they are believed to capture richer information about a query's intent and difficulty.
However, our experiments that compare query embeddings from Qwen2.5 models~\citep{qwen25} ranging from 0.5B to 72B parameters find little improvement in routing performance as model size increases (Figure~\ref{fig:encoder-scaling}b, Appendix~\ref{app:scaling}).
This suggests that the query properties required by routing might already have been captured by small encoders.
We therefore investigate which query properties are critical for routing and whether they can be extracted directly from text without a neural text encoder.

To identify these query properties, we use sparse autoencoders (SAEs)~\citep{bricken2023monosemanticity,cunningham2023sparseautoencodershighlyinterpretable} to decompose LM (Language Model) hidden states into sparse combinations of interpretable latent features.
The latent descriptions suggest which text properties to measure, while their activation patterns provide targets for text extractors to approximate.
However, computing these activations still requires running both the LM and the SAE.
We therefore propose \ours{}, a pipeline that constructs regular-expression (regex) feature extractors to approximate the activation patterns of related SAE latents directly from text.
During offline discovery, an LLM uses descriptions of grouped latents to author regex feature extractors, then refines them using feedback on how well their outputs align with latents' activations.
The resulting \emph{128} fixed regex extractors compute interpretable query features for routing without an LM or SAE forward pass at inference.

With just \emph{128} regex features, \ours{} achieves \emph{76.43\%} average routing accuracy across four benchmarks, comparable to \emph{74.92--76.41\%} for routers using neural text encoders.
\ours{} offers three additional benefits.
First, \emph{interpretability}: each regex feature measures a meaningful property directly from the query string.
Second, \emph{low latency}: our router runs on CPU with nearly 90\% lower median routing latency than the Qwen2.5-0.5B router on GPU.
Third, \emph{robustness}: our router achieves the highest average routing accuracy under four settings of query perturbation.

We make three contributions:
\begin{enumerate}
    \item We propose \ours{}, an SAE-guided pipeline that discovers and refines interpretable regex feature extractors from unlabeled text.
    \item We show that one fixed set of 128 regex features can replace neural text encoders while achieving comparable routing accuracy across four benchmarks.
    \item Our findings open a new direction for the community to design and understand LLM routers through explicit, interpretable text features.
\end{enumerate}

\section{Related work}
\label{sec:related}

\paragraph{LLM routing.}
Difficulty-aware routers estimate how hard a query is and select the most capable model to handle it \citep{zhuang2024embedllmlearningcompactrepresentations, feng2025graphroutergraphbasedrouterllm, song2025irtroutereffectiveinterpretablemultillm, wang2025iclrouterincontextlearnedmodel, ding2025bestrouteadaptivellmrouting}.
Preference-aligned routers use pairwise human preference \citep{ong2025routellmlearningroutellms, frick2025prompttoleaderboard}.
Clustering-based routers assign each query cluster its most cost-effective model \citep{jitkrittum2025universalmodelroutingefficient, zhang2025beyond}.
These approaches all use a neural text encoder to represent a query.
Interpretable routers use predefined query concepts or score generated tags from model-performance labels~\citep{storek2026routesplainfaithfulintervenablerouting,chen2025tagrouterlearningroutellms}.
Whereas our regex features are discovered from unlabeled text, without routing labels or a predefined task schema.

\paragraph{SAEs and interpretable text representations.}
Sparse autoencoders (SAEs) decompose language-model activations into sparse combinations of interpretable latent features~\citep{cunningham2023sparseautoencodershighlyinterpretable,bricken2023monosemanticity}.
Recent work has scaled SAEs and released feature dictionaries for several model families~\citep{gao2024scalingevaluatingsparseautoencoders,lieberum2024gemmascopeopensparse,he2024llamascopeextractingmillions,deng2026qwenscopeturningsparsefeatures}.
LMs can describe these latents from their top-activating contexts~\citep{bills2023language,paulo2025automaticallyinterpretingmillionsfeatures}.
A complementary line represents text through a small set of named properties.
These properties may be specified from linguistic theory~\citep{zhou2024llmfeaturebasedframeworkdialogue}, proposed by an LM conditioned on class names or a target variable~\citep{sun2025conceptbottlenecklargelanguage,Balek_2025,benara2024craftinginterpretableembeddingsasking}, or selected using labeled downstream data~\citep{ludan2024interpretablebydesigntextunderstandingiteratively,feng2025bayesianconceptbottleneckmodels}.
Their values are typically computed by prompting an LM or by probing a neural embedding at inference time.
\ours{} uses SAE latents to discover regex features computed directly from text without neural encoding.

\section{Preliminaries}
\label{sec:prelim}

\paragraph{LLM routing.}
Given a query $q$ and a pool of candidate language models $\mathcal{L}=\{\ell_1,\ldots,\ell_M\}$, a router selects the optimal model to answer the query.
To train and evaluate a router, we use a dataset $\mathcal{D}_{\mathrm{route}}=\{(q_i,\mathbf{r}_i)\}_{i=1}^{n}$, where $\mathbf{r}_i\in\mathbb{R}^{M}$ records the scores of the candidate models' responses to $q_i$.
We divide this dataset into disjoint training, validation, and test splits, denoted by $\mathcal{D}_{\mathrm{route}}^{\mathrm{train}}$, $\mathcal{D}_{\mathrm{route}}^{\mathrm{val}}$, and $\mathcal{D}_{\mathrm{route}}^{\mathrm{test}}$.
These splits are used to fit the router, select its hyperparameters, and evaluate its performance, respectively.
For a router $\pi$ that maps a query to a model index, its test performance is
\begin{equation}
    R\!\left(\pi;\mathcal{D}_{\mathrm{route}}^{\mathrm{test}}\right)
    =\frac{1}{|\mathcal{D}_{\mathrm{route}}^{\mathrm{test}}|}
    \sum_{(q_i,\mathbf{r}_i)\in\mathcal{D}_{\mathrm{route}}^{\mathrm{test}}}
    r_{i,\pi(q_i)}.
    \label{eq:routing-performance}
\end{equation}

\paragraph{Unlabeled discovery text.}
To keep feature discovery independent of the routing benchmarks, we use a general \textbf{unlabeled web text corpus} $\mathcal{D}_{\mathrm{text}}$, without access to routing benchmark data or evaluation results.
We partition it into disjoint fit, $\mathcal{D}_{\mathrm{text}}^{\mathrm{fit}}$, and held-out portions, $\mathcal{D}_{\mathrm{text}}^{\mathrm{held}}$.
We use $\mathcal{D}_{\mathrm{text}}^{\mathrm{fit}}$ to iteratively refine and select the regex feature extractors, then freeze them and evaluate their alignment with SAE activations on $\mathcal{D}_{\mathrm{text}}^{\mathrm{held}}$.

\paragraph{SAE latents and activations.}
Given a language-model hidden activation $\mathbf{h}_{i,t}\in\mathbb{R}^{d}$ at token position $t$ in $x_i\in\mathcal{D}_{\mathrm{text}}$, an SAE encoder $E_{\mathrm{SAE}}$ maps it to a sparse vector of latent activations:
\begin{equation}
    \mathbf{z}_{i,t}
    = E_{\mathrm{SAE}}(\mathbf{h}_{i,t})\in\mathbb{R}_{\geq 0}^{K}.
    \label{eq:sae-encoding-prelim}
\end{equation}
Here, $K$ is the number of latent features, and $z_{i,t,k}$ measures the activation of latent $k$ at token position $t$.
Each latent is associated with a human-readable description from Neuronpedia~\citep{neuronpedia}.
Summing over the token positions $\mathcal{T}_i$ of document $x_i$ gives latent $k$'s document-level activation:
\begin{equation}
    s_{i,k}=\sum_{t\in\mathcal{T}_i}z_{i,t,k}.
    \label{eq:document-latent-prelim}
\end{equation}

\section{Method}
\label{sec:method}

\begin{figure}[t]
    \centering
    \includegraphics[width=\linewidth]{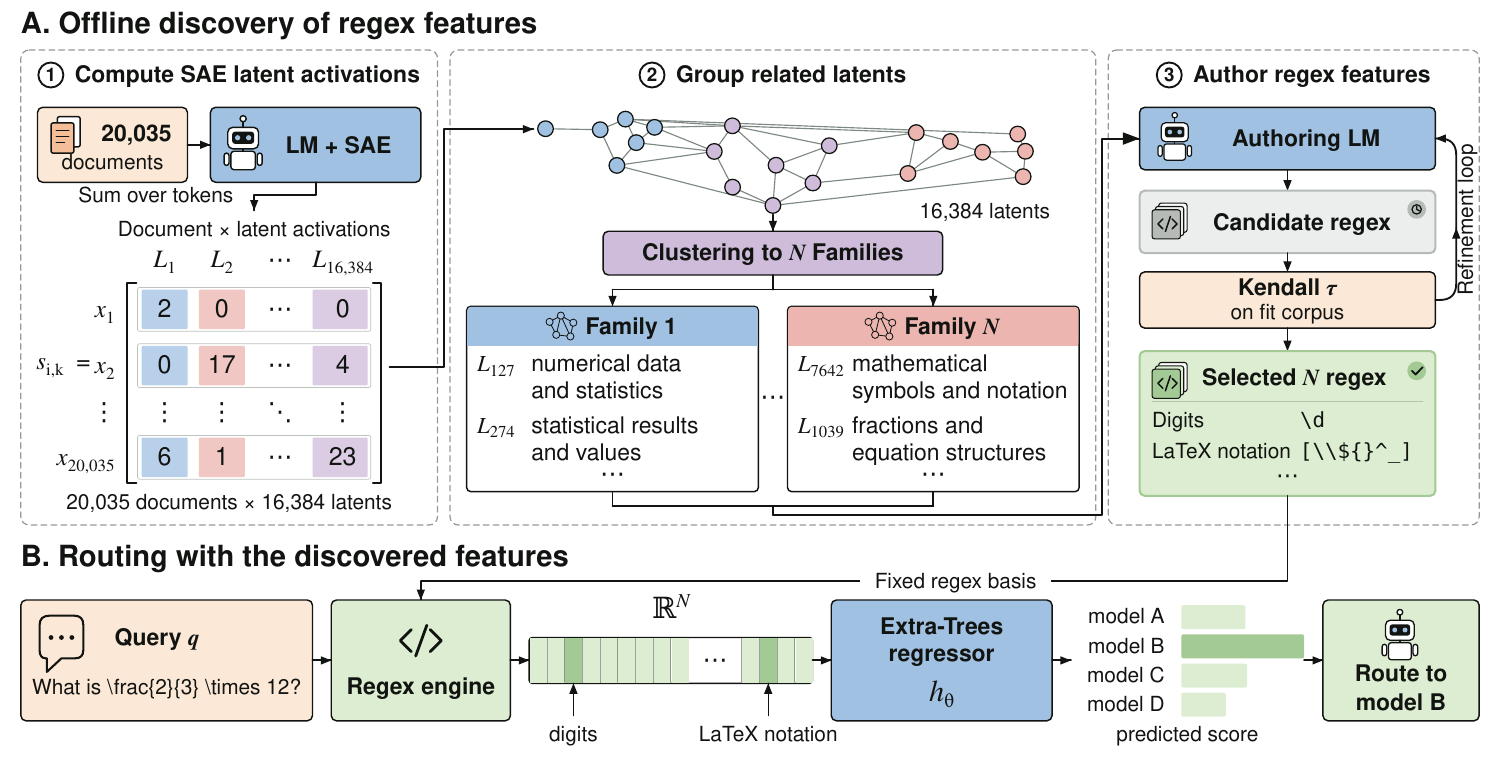}
    \caption{Overview of our pipeline. \textbf{Top}: offline discovery uses a language model and SAE to organize latent families and author regex feature extractors from public text. \textbf{Bottom}: at inference, the fixed extractors map a query to a feature vector which are used to select a model.}
    \label{fig1}
\end{figure}

Our goal is to route queries using a compact set of interpretable features extracted directly from text.
To discover such features, \ours{} uses SAE descriptions and activations to construct regex feature extractors (Figure~\ref{fig1}, top).
Each extractor combines a regex with an aggregation rule that outputs a numerical feature value.
At inference, the router computes feature values and selects a model without running an LM or SAE (Figure~\ref{fig1}, bottom).

\subsection{Discovering regex feature extractors}
\label{sec:method-discover}

\paragraph{Grouping SAE latents.}

To construct regex feature extractors, we determine the latent properties they will approximate.
Constructing and refining an extractor for each of the 16,384 SAE latents would be costly and require thousands of extractor evaluations per query.
Moreover, related properties can be distributed across several specialized latents~\citep{bricken2023monosemanticity,chanin2024absorption,leask2025canonical}, so a single latent may be too narrow for a reusable text feature.
We therefore group related latents into \emph{latent families} and construct one extractor per family.

We group latents using the statistical association between their document-level activations $s_{i,k}$, defined in Equation~\ref{eq:document-latent-prelim}.
Specifically, we define their affinity as the magnitude of the correlation across documents:
\begin{equation}
    C_{k,k'}=\left|\operatorname{corr}\!\left(\mathbf{s}_{:,k},\mathbf{s}_{:,k'}\right)\right|.
    \label{eq:grouping-affinity-revised}
\end{equation}

We compute the affinity for every pair of latents and cluster them into $N$ families, $\mathcal{F}_1,\ldots,\mathcal{F}_N$.
We use normalized spectral clustering~\citep{ng2001spectral}, which produces a smaller maximum family size among the grouping methods evaluated in Section~\ref{sec:ablation-grouping}.
This helps limit the number of latent properties assigned to a single extractor.

\paragraph{Authoring regex feature extractors.}
\label{sec:method-author}

For each latent family, we aim to write a regex extractor that approximates the activations of its member latents.
To author such an executable text pattern, we prompt an LLM with descriptions of the member latents from Neuronpedia~\citep{neuronpedia} and tokens associated with the family’s activations. 
The descriptions suggest what the family captures, while the associated tokens provide examples of how that property appears in text. 
We then ask an LLM to identify reusable surface patterns, such as punctuation, capitalization, and formal notation, rather than copy topical vocabulary lists.
The LLM initially proposes three candidate extractors per family, each with a human-readable feature name $n_j$, a regex $p_j$, and an aggregation rule $a_j$ that converts matches into a scalar, such as their count or proportion in the document.
The resulting feature is $\phi_j(x)=a_j(\mathcal{M}_{p_j}(x),x)$, where $\mathcal{M}_{p_j}(x)$ collects the pattern's matches.
The initial authoring prompt is provided in Appendix~\ref{app:regex-authoring-template}, with an illustrated example in Figure~\ref{fig:regex-authoring-example}.

\paragraph{Refining regexes through activation alignment.}
\label{sec:method-refine}

To better capture each latent family's behavior, we refine the regex extractors using alignment between their outputs and family activations.
We compare feature values extracted by each candidate for document $x_i$ with a reference value derived from the activations of latent family $\mathcal{F}_j$:
\begin{equation}
    y_{i,j}
    =\sum_{k\in\mathcal{F}_j}\sum_{t\in\mathcal{T}_i}
      \mathbf{1}\!\left[z_{i,t,k}>0\right].
    \label{eq:family-target}
\end{equation}
The indicator reflects how often the family's latents are active throughout the document.
We measure the alignment between regex output $\phi_j(x_i)$ and the reference $y_{i,j}$ using Kendall's $\tau_b$,
\begin{equation}
    J_j(p_j,a_j)
    =\tau_b\!\left(
      \{\phi_j(x_i)\},
      \{y_{i,j}\}
    \right), {x_i\in\mathcal{D}_{\mathrm{text}}^{\mathrm{fit}}}.
    \label{eq:regex-alignment}
\end{equation}

Higher alignment means that the regex more faithfully preserves the ordering of documents.
We return the candidate regexes and their fit alignment scores to the LLM, which proposes revised or alternative extractors.
We select the candidate with the highest fit alignment for each family and freeze the resulting $N$ extractors.
The refinement prompt is provided in Appendix~\ref{app:regex-refinement-details}.

\subsection{Routing with the discovered regex extractors}
\label{sec:method-use}

After freezing the regex library, we train only the routing head on $\mathcal{D}_{\mathrm{route}}^{\mathrm{train}}$, using the same extractors across all benchmarks.
For each query $q_i$, we apply the $N$ frozen regex feature extractors $(p_j,a_j)$ to obtain the feature vector (Figure~\ref{fig1}, bottom):
\begin{equation}
    \boldsymbol{\phi}(q_i)
    =\left[\phi_1(q_i),\ldots,\phi_N(q_i)\right]\in\mathbb{R}^{N}.
    \label{eq:regex-representation}
\end{equation}

We then fit a multi-output regressor taking $\boldsymbol{\phi}(q_i)$ as input:
\begin{equation}
    h_{\theta}:\mathbb{R}^{N}\rightarrow\mathbb{R}^{M}
\end{equation}
to predict the response-score vector $\mathbf{r}_i$.
Our primary multi-output regressor is an Extra-Trees regressor~\citep{geurts2006extremely}, trained by minimizing prediction error across candidate models,
\begin{equation}
    \widehat{\theta}
    =\arg\min_{\theta}
      \sum_{(q_i,\mathbf{r}_i)\in\mathcal{D}_{\mathrm{route}}^{\mathrm{train}}}
      \left\|
        h_{\theta}\!\left(\boldsymbol{\phi}(q_i)\right)
        -\mathbf{r}_i
      \right\|_2^2.
    \label{eq:reader-objective}
\end{equation}
At inference, we extract regex features from query $q$ and use the trained regressor to select the model with the highest predicted score, without an LM or SAE forward pass.

\section{Experiments}
\label{sec:experiments}

\subsection{Experimental setup}
\label{sec:setup}

\paragraph{Benchmarks.}
We conduct experiments on four benchmarks: EmbedLLM~\citep{zhuang2024embedllmlearningcompactrepresentations}, NineBy30k~\citep{lu2026routingplateau}, R2Bench~\citep{xue2026r2router}, and CARROT~\citep{somerstep2025carrot}.
Table~\ref{tab:benchmarks} summarizes their candidate pools and the splits shared by all methods.
We remove prompt templates specific to each source dataset as detailed in Appendix~\ref{app:preprocessing}. 

\paragraph{Pipeline configuration.}
The discovery corpus $\mathcal{D}_{\mathrm{text}}$ contains 20,035 public documents from FineWeb~\citep{penedo2024fineweb}, OpenWebMath~\citep{paster2023openwebmath}, CodeParrot, Alpaca~\citep{taori2023alpaca}, and OpenAssistant~\citep{kopf2023openassistant}, covering web prose, mathematics, code, and instruction/dialogue text.
The fit and held-out portions contain 13,394 and 6,641 documents, respectively.
Our default configuration uses Gemma-2-2B~\citep{gemma2} and its 16,384-latent Gemma Scope SAE~\citep{lieberum2024gemmascopeopensparse} at the final transformer layer.
We group latents into $N=128$ families and use Claude Opus 5 Medium to author and refine regex extractors.
Appendix~\ref{app:pipeline-config} provides the corpus sources and detailed pipeline configuration.

\paragraph{Baselines.}
We compare query embeddings produced by three types of neural text encoders.
\textbf{Sentence-embedding models} are trained for semantic similarity and retrieval. Our baselines include MiniLM-L6~\citep{minilm}, MPNet-base~\citep{mpnet}, BGE-base~\citep{bge}, GTE-base~\citep{gte}, and E5-base~\citep{e5}.
\textbf{Encoder-only LMs} are pretrained through masked-token prediction. We evaluate ModernBERT-base and ModernBERT-large~\citep{modernbert}.
\textbf{Decoder-only LMs}, pretrained to predict the next token, include Qwen2.5-0.5B~\citep{qwen25}, Gemma-2-2B~\citep{gemma2}, and Llama-3.1-8B~\citep{llama3}.
We also include TF-IDF as a baseline without a neural encoder.
Baseline details are in Appendix~\ref{app:encoders}.

\paragraph{Evaluation.}
For each representation and benchmark, we train an Extra-Trees~\citep{geurts2006extremely} routing head and select its hyperparameters on the validation data (Appendix~\ref{app:heads}).
We report routing accuracy (equation~\ref{eq:routing-performance}), for each benchmark and its unweighted average across the four benchmarks.

\begin{table}[!htbp]
\centering
\small
\setlength{\tabcolsep}{5pt}
\caption{\textbf{Test routing accuracy (\%).} All routers use Extra-Trees heads with hyperparameters selected on validation. Column maxima are in \textbf{bold}.}
\label{tab:main}
\begin{tabular}{@{}lccccc@{}}
\toprule
Representation & EmbedLLM & NineBy30k & R2Bench & CARROT & \textsc{Avg.} \\
\midrule
Best single model & 60.73 & 67.34 & 81.10 & 84.82 & 73.50 \\
\midrule
\multicolumn{6}{@{}l}{\textbf{Sentence embeddings}} \\
\addlinespace[2pt]
MiniLM-L6~\citep{minilm} & 64.87 & 67.47 & 81.56 & 86.18 & 75.02 \\
MPNet-base~\citep{mpnet} & 64.73 & 67.43 & 81.39 & 86.28 & 74.96 \\
BGE-base~\citep{bge}     & 64.47 & 67.77 & 81.53 & 86.51 & 75.07 \\
GTE-base~\citep{gte}     & 64.37 & 67.63 & 81.77 & 86.60 & 75.09 \\
E5-base~\citep{e5}       & 63.77 & 67.59 & 81.86 & 86.47 & 74.92 \\
\midrule
\multicolumn{6}{@{}l}{\textbf{Embeddings from encoder-only LMs}} \\
\addlinespace[2pt]
ModernBERT-base~\citep{modernbert}  & 65.60 & 67.88 & 81.94 & 86.88 & 75.58 \\
ModernBERT-large~\citep{modernbert} & 66.27 & 68.07 & 82.19 & 87.10 & 75.91 \\
\midrule
\multicolumn{6}{@{}l}{\textbf{Embeddings from decoder-only LMs}} \\
\addlinespace[2pt]
Qwen2.5-0.5B~\citep{qwen25}          & 67.50 & 68.23 & 82.38 & \textbf{87.20} & 76.33 \\
Gemma-2-2B~\citep{gemma2}            & \textbf{67.83} & 68.26 & 82.38 & 87.08 & 76.39 \\
Llama-3.1-8B~\citep{llama3}          & 67.77 & 68.13 & \textbf{82.70} & 87.02 & 76.41 \\
\midrule
\multicolumn{6}{@{}l}{\textbf{Non-neural features (without embedding)}} \\
\addlinespace[2pt]
TF-IDF~\citep{salton1988termweighting} & 65.60 & 68.23 & 81.99 & 86.99 & 75.70 \\
Ours ($N=32$)  & 66.63 & 68.50 & 82.15 & 86.96 & 76.06 \\
Ours ($N=128$) & 67.37 & \textbf{69.01} & 82.16 & 87.18 & \textbf{76.43} \\
\bottomrule
\end{tabular}
\end{table}

\subsection{Routing quality without embeddings}
\label{sec:main}
\begin{figure}[!htbp]
\centering
\begin{minipage}{0.85\linewidth}
\begin{subfigure}[t]{0.49\linewidth}
\centering
\includegraphics[width=\linewidth]{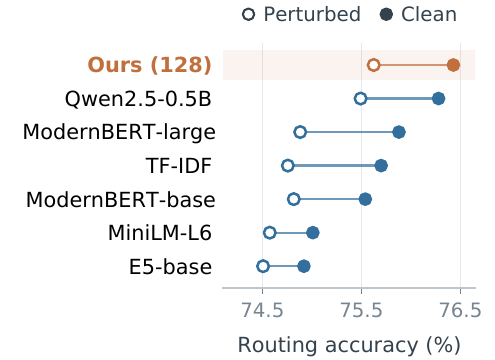}
\caption{Robustness}
\label{fig:robust}
\end{subfigure}\hfill
\begin{subfigure}[t]{0.49\linewidth}
\centering
\includegraphics[width=\linewidth]{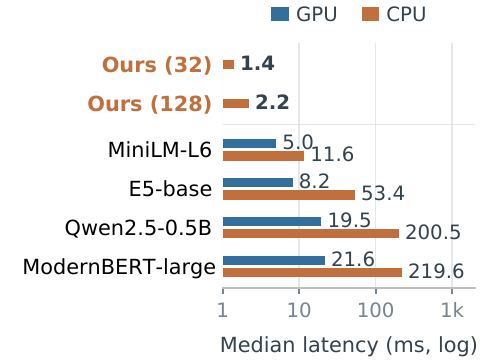}
\caption{Routing latency}
\label{fig:routing-latency}
\end{subfigure}
\end{minipage}
\caption{Robustness and routing latency. Our regex router retains the highest perturbed accuracy (left) and achieves lower median latency on CPU than neural encoder routers on GPU (right).}
\label{fig:robust-latency}
\end{figure}

Our 128 regex features achieve 76.43\% mean routing accuracy, comparable to the 76.41\% achieved by the strongest embedding baseline (Table~\ref{tab:main}).
With only 32 regex features, our router achieves 76.06\% mean routing accuracy, exceeding the mean accuracy of all sentence-embedding and encoder-only LM baselines in this comparison.
Both configurations also outperform TF-IDF (75.70\%), suggesting that our regex features provide more effective routing signals than word- and character-$n$-gram frequencies alone.
These results show that a compact set of surface-extractable features can replace a neural text encoder while retaining comparable average routing accuracy.

\subsection{Robustness to query perturbations}
\label{sec:robustness}

Because regex features depend on surface-text patterns, we test whether changes in wording or formatting undermine routing reliability.
Following RouterArena~\citep{lu2025routerarena}, we apply four types of query perturbation: misspellings, word substitutions, sentence restructuring, and paraphrasing.
All methods are fitted on clean text and evaluated on perturbed queries using the original response scores.
Figure~\ref{fig:robust-latency} (left) reports clean accuracy and mean accuracy of four perturbation rewrites. 
Our router retains the highest mean perturbed accuracy among the evaluated representations (75.62\%), with a drop in accuracy similar to that of Qwen2.5-0.5B.
Sentence-embedding models lose less accuracy but still have lower perturbed scores.
This suggests that \ours{} features remain useful even when query wording and format change.
The complete setup and results are in Appendix~\ref{app:robustness}.

\subsection{Routing latency}
\label{sec:efficiency}

From Figure~\ref{fig:robust-latency} (right), \ours{} preserves routing quality at lower computational cost.
We measure routing latency from query input to model selection, including feature extraction and routing-head inference, which adds to the time to first token (TTFT). 
With 128 features, our router's median latency is 2.2\,ms on CPU, compared with 19.5\,ms for Qwen2.5-0.5B on GPU.
Running the router on the CPU leaves GPU resources for response generation and can simplify deployment on edge devices with limited access to accelerators.
The full table is in Appendix~\ref{app:latency}.

\FloatBarrier
\section{Ablation Studies}
\label{sec:ablations}

We conduct ablations to understand the contributions of each component of our pipeline.
Unless otherwise stated, we follow the pipeline configuration and evaluation protocol in Section~\ref{sec:setup}.

\begin{figure}[!htb]
\centering
\includegraphics[width=0.9\linewidth]{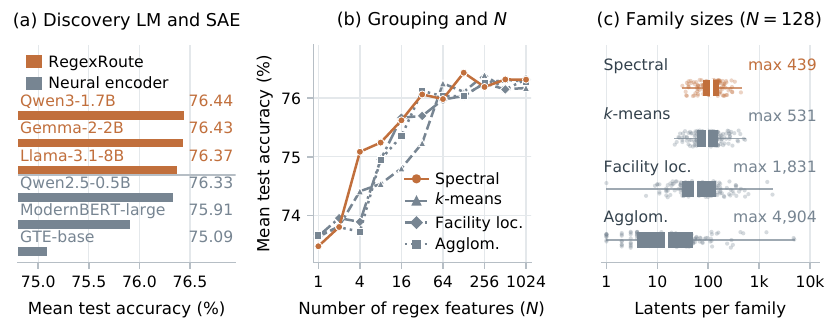}
\caption{\textbf{Pipeline ablations.} Mean routing accuracy across four benchmarks for (a) discovery LM--SAE pairs at $N=128$ and (b) latent grouping strategy across feature counts, $N$. (c) Latent family sizes at $N=128$ by different grouping strategies.}
\label{fig:ablation-overview}
\end{figure}

\subsection{Choice of discovery LM and SAE}
\label{sec:ablation-source}

We test whether \ours{} depends on a particular family of discovery models.
Alongside the default Gemma configuration, we use Qwen3-1.7B and Llama-3.1-8B with their pretrained SAEs~\citep{deng2026qwenscopeturningsparsefeatures,he2024llamascopeextractingmillions} to generate 128 regex feature extractors per model with the \ours{} pipeline (Figure~\ref{fig1}).
In Figure~\ref{fig:ablation-overview}(a), Qwen and Llama reach 76.44\% and 76.37\%, respectively, compared with 76.43\% for the main Gemma configuration.
\ours{} is not specific to a single base model or SAE and can produce general regex feature extractors from multiple LM-SAE pairs.
Per-benchmark results and SAE dictionary sizes are provided in Appendix~\ref{app:discovery-scaling-results}.

\subsection{Number of regex features}
\label{sec:ablation-families}

For each feature count N, we evaluate routing performance using a new regex library authored by the LLM.
The orange curve in Figure~\ref{fig:ablation-overview}(b) shows that accuracy increases by 2.6~pp from $N=1$ to $32$, as additional regex features capture a broader range of routing-important surface patterns.
Then, performance saturates between $N=32$ and $128$, suggesting that a compact feature set captures most of the important feature directions, while further expansion primarily increases extraction cost.
We retain $N=128$ as the main configuration, with $N=32$ providing a lower-latency alternative.

\subsection{Choice of latent grouping}
\label{sec:ablation-grouping}

We compare spectral clustering with spherical $k$-means, facility-location selection, and average-linkage agglomerative clustering.
For $N\geq64$, mean routing accuracy differs by at most 0.40 percentage points across the four methods (Figure~\ref{fig:ablation-overview}b).
Spectral clustering performs best at $N=4$ and $8$.
At $N=128$, its largest family contains 439 latents, compared with 531 to 4{,}904 for the alternatives (Figure~\ref{fig:ablation-overview}c).
We retain spectral clustering to limit the maximum number of latents each regex extractor must approximate while maintaining competitive routing accuracy.
Appendix~\ref{app:grouping} provides detailed results and descriptions of the grouping methods.

\subsection{Do SAE guidance and refinement improve regex features for routing?}
\label{sec:ablation-authoring}

We test whether SAE-guided authoring and activation-alignment refinement (Sec~\ref{sec:method-discover}) improve the regex features used for routing.
At $N\in\{32,128\}$, we compare our full procedure with two ablations: the \emph{LLM-only} variant, which omits both SAE guidance and refinement, and the \emph{initial regexes} variant, which uses our pipeline without the final refinement stage.
The LLM-only authoring prompt is provided in Appendix~\ref{app:regex-control-template}, and experimental details are provided in Appendix~\ref{app:authoring-ablations}.

\begin{table}[!htb]
\centering
\small
\setlength{\tabcolsep}{4pt}
\caption{\textbf{Ablating SAE guidance and refinement.} Checkmarks indicate enabled components. Held alignment is Kendall $\tau_b$ from equation~\ref{eq:regex-alignment}. The test is a four-benchmark mean routing accuracy (\%).}
\label{tab:authoring-ablation}
\begin{tabular}{lcccccc}
\toprule
 & & & \multicolumn{2}{c}{$N=32$} & \multicolumn{2}{c}{$N=128$} \\
\cmidrule(lr){4-5}\cmidrule(lr){6-7}
Procedure & \shortstack{SAE\\guidance} & Refinement & \shortstack{Held\\alignment} & Test (\%) & \shortstack{Held\\alignment} & Test (\%) \\
\midrule
LLM-only & --- & --- & 0.2343 & 74.68 & 0.2852 & 75.67 \\
Initial regexes & $\checkmark$ & --- & 0.3770 & 75.68 & 0.3865 & 76.16 \\
\textbf{Ours} & $\checkmark$ & $\checkmark$ & \textbf{0.4470} & \textbf{76.06} & \textbf{0.4709} & \textbf{76.43} \\
\bottomrule
\end{tabular}

\end{table}

\paragraph{SAE-guided authoring improves over the LLM-only control.}
Comparing \emph{initial regexes} with \emph{LLM-only} in Table~\ref{tab:authoring-ablation}, we find that SAE-guided authoring raises test accuracy by 1.00 and 0.49 percentage points at $N=32$ and $N=128$, respectively.
Latent descriptions identify properties for the LLM to capture, while associated tokens illustrate how they appear in text.
The gains suggest that these concrete targets help guide the LLM toward regex features useful for routing.
\paragraph{Alignment-based refinement further improves routing.}
Higher alignment in Equation~\ref{eq:regex-alignment} means that an extractor more consistently assigns larger values to documents where its SAE family activates more frequently.
Refinement increases test routing accuracy by 0.38 and 0.27 percentage points at $N=32$ and $128$, respectively.
These gains suggest that more faithfully approximating SAE activation patterns helps regex features capture information useful for routing.

\subsection{Choice of routing head}
\label{sec:ablation-heads}

We ablate five routing heads, with setup and full results in Appendix~\ref{app:heads}.
In the recorded results, regex features are less performant with $k$-NN and $k$-means, but achieve the highest mean accuracy with tree-based heads.
This is consistent with the structure of our representation: its coordinates measure distinct properties through counts and ratios, and a fixed distance across them may not capture the query differences.
Tree-based heads instead learn thresholds for individual features from routing labels, allowing them to partition queries based on properties relevant to model selection.

\begin{figure}[!b]
  \centering
  \includegraphics[width=1\linewidth]{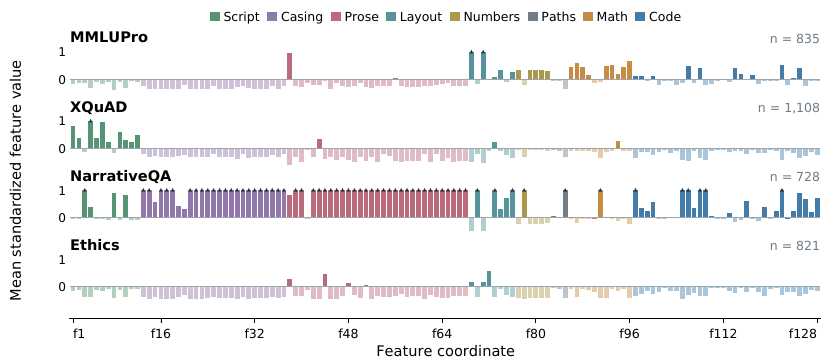}
  \caption{\textbf{Different query sources peak at different feature coordinates.} Values above $+1$ standard deviation are capped and marked with triangles.}
  \label{fig:source-feature-centers}
\end{figure}

\section{Analysis}
\label{sec:analysis}
\vspace{-0.5em}
We aim to investigate why surface features achieve routing accuracy comparable to that of dense embeddings and why larger encoders yield limited gains. We provide the complete set of 128 regex feature extractors for reference in Appendix~\ref{app:regex-features}.

\subsection{Interpretable surface features distinguish queries}
\label{sec:interp}
\vspace{-0.5em}

\paragraph{Readable differences between query sources.}
Using source-dataset labels only for post-hoc analysis, Figure~\ref{fig:source-feature-centers} shows that the independently discovered features expose recognizable differences in query properties across datasets.
We group the 128 features into eight categories of surface properties: script, casing, prose, layout, numbers, paths, math, and code.
For example, MMLUPro is elevated on short-line and numeric features, XQuAD on script and accent features, and NarrativeQA on prose word counts.
These patterns provide interpretable cues for distinguishing query sources.
We include additional profiles in Appendix~\ref{app:source-profiles}.

\paragraph{Regex features group queries by source and model performance.}

\begin{figure}[t]
  \centering
  \includegraphics[width=0.85\linewidth]{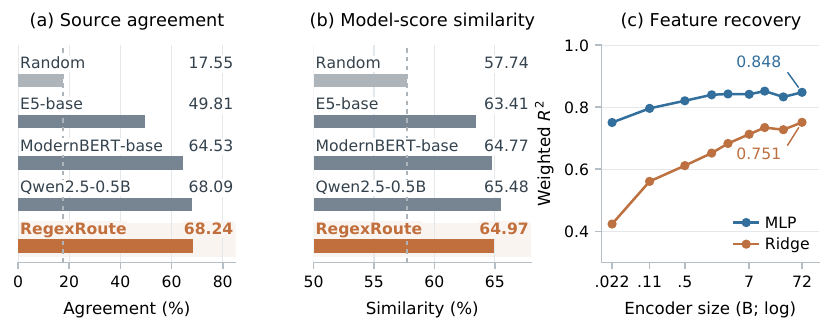}
  \caption{(a,b) Source agreement and model-score similarity. (c) Feature-recovery $R^2$ across neural text encoder sizes.}
  \label{fig:query-neighborhoods}
  \label{fig:feature-recovery-scaling}
\end{figure}

We measure whether neighboring queries tend to share a similar source dataset and observed model score.
For each test query, we select $k=20$ other test queries that share the most leaves with it across the fitted Extra-Trees router.
Figure~\ref{fig:query-neighborhoods}(a,b) shows that regex features group queries by source and model performance about as well as Qwen2.5-0.5B and better than E5-base and ModernBERT-base.
Both measures significantly exceed the random baseline, suggesting that surface patterns provide useful cues for routing without neural encoding.
Appendix~\ref{app:query-neighborhoods} provides additional settings and results.

\subsection{Small encoders already capture surface features}
\label{sec:feature-recovery}

To understand the limited routing gains from encoder scaling, we examine whether larger encoders better capture query properties useful for routing.
Using our 128 regex features as concrete targets, we train nonlinear MLP and linear ridge probes on each benchmark's training split to predict their values from frozen embeddings.
We evaluate feature recovery on held-out test queries using $R^2$ and report the average across four benchmarks.
Higher $R^2$ indicates that these surface properties are more accurately recoverable from the encoder's embeddings, with $R^2=1$ denoting perfect recovery.
Appendix~\ref{app:feature-recovery} provides the detailed setup and per-benchmark results.

With an MLP probe, Qwen2.5-0.5B reaches $R^2=0.820$, whereas the roughly $147\times$ larger 72B model only reaches 0.848 (Figure~\ref{fig:feature-recovery-scaling}(c)).
Linear recovery improves more substantially, from 0.612 to 0.751.
The smaller gain with the MLP probe suggests that much of this information is already recoverable from the small encoder through a nonlinear mapping.
These results show that a small encoder already captures surface features well in its dense embeddings.
Because these features describe patterns directly observable in the input text, even a small encoder could recognize them without complex semantic reasoning.
Together, these findings suggest that small encoders already capture much of the surface information sufficient for competitive routing, helping explain why larger encoders yield limited gains on these benchmarks.

\section{Conclusion}
\label{sec:conclusion}

We presented \ours{}, which uses SAE descriptions and activation patterns to discover interpretable regex feature extractors from unlabeled text.
A fixed library of 128 regex extractors achieves routing accuracy comparable to neural embeddings across four benchmarks, with low CPU latency and competitive performance under query perturbations.
Our ablations show that SAE guidance and activation-based refinement improve routing performance, and that the pipeline works across the evaluated discovery model families.
Our analysis connects these features to query source and model performance and shows that small encoders already represent them well.
Together, these findings establish that a compact set of explicit surface features is sufficient for competitive routing, offering a possible explanation to the limited benefit of larger encoders.

\bibliography{iclr2027_conference}
\bibliographystyle{iclr2027_conference}

\appendix
\section{Prompts for Regex Authoring}
\label{app:regex-authoring-prompt}

We provide three prompt templates for authoring regex feature extractors: SAE-guided initial authoring, alignment-based refinement, and authoring without SAE guidance.
For the ablation in Section~\ref{sec:ablation-authoring}, each variant proposes six candidates per output feature in two rounds of three, then retains one.
Duplicate proposals count toward this budget, and all features complete both rounds without early stopping.

\begin{figure}[htbp]
  \centering
  \includegraphics[width=\linewidth]{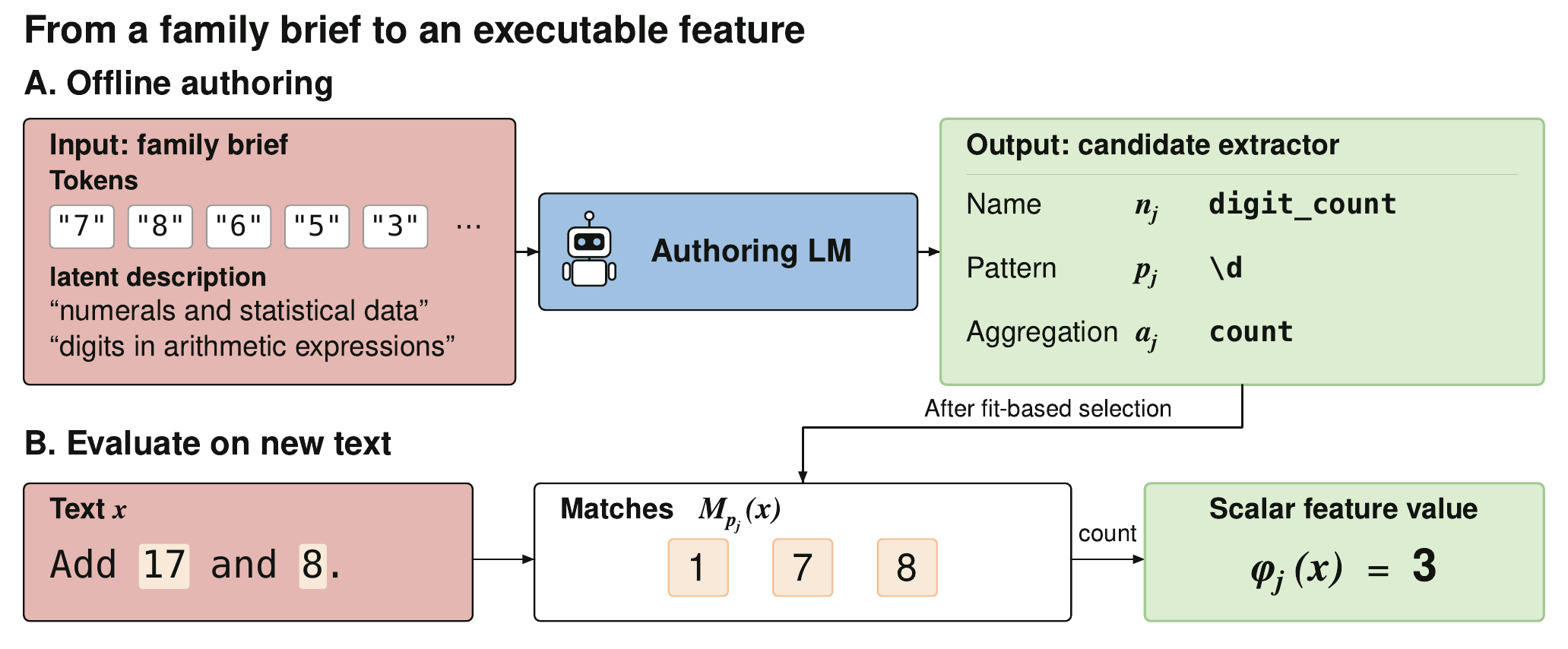}
  \caption{Illustrative regex-authoring example. Member-latent descriptions and associated tokens inform a candidate named \texttt{digit\_count}, with pattern $p_j=\texttt{\textbackslash d}$ and count aggregation $a_j$. Once selected and frozen, the extractor counts the matches \texttt{1}, \texttt{7}, and \texttt{8} in \texttt{Add 17 and 8.}, giving $\phi_j(x)=3$.}
  \label{fig:regex-authoring-example}
\end{figure}

\subsection{SAE-guided initial authoring}
\label{app:regex-authoring-template}

This prompt is used for the initial authoring stage in Section~\ref{sec:method-discover}, proposing regex feature extractors for each SAE latent family using its descriptions and associated tokens.
For the variant without refinement, we use this template for both rounds, supplying the first three candidates in round two without alignment scores or other fit feedback.
The controller then selects among all six candidates using fit-set alignment.

\begingroup
\small
\begin{verbatim}
You are authoring regex feature extractors for latent families of a
sparse autoencoder. Work only from the supplied family briefs and
previous candidates. All family evidence comes from the unlabeled
corpus.
No routing benchmark information is available. Do not use tools.
An external controller performs scoring and retention; return only
the candidates requested for this round.

INPUT
Batch {batch_id} of six; library width {N}; round {round_id}.
<family_brief>
{fid, number of latents, activation share,
  Neuronpedia descriptions, highly associated tokens,
  family-specific tokens, punctuation/symbol tokens,
  digit-bearing tokens, for each assigned family}
</family_brief>
<round_request>
{"candidates_per_active_fid": {each assigned fid: 3},
 "previous_candidates": [
   {fid, name, pattern, agg, flags} for each previous candidate],
 "previous_fit_scores": []}
</round_request>
previous_candidates is empty in round 1. In round 2, it contains
the first three candidates per family, without evaluation feedback.

TASK
Propose three diverse extractors per family whose values track the
ranking of its member-latent firing-event count across documents.
Candidates are scored by Kendall tau-b on fit documents.
In round 2, propose three additional candidates; do not repeat
the previous candidate array.
If brief tokens are content words, capture their shared form,
not the words themselves.

CONSTRAINTS
Use symbols, orthography, and structure: punctuation, digits,
operators, brackets, casing, whitespace/layout, Unicode classes,
run lengths, and closed-class code/LaTeX/HTML/SQL keywords.
Do not use English content-word literals, topical vocabularies,
or morphological suffix patterns. Use non-capturing groups (?:...).

AGGREGATORS
count: number of matches; ratio: matches per approximate word;
binary: whether any match exists; maxrun: longest match length.
Flags: "", "i", "m", or "im".

Prefer ratio when fit scores are within 0.01 of the best;
within this margin, prefer count/ratio over binary/maxrun.

OUTPUT
Return only a JSON array, without prose or markdown. Each object
must contain fid (integer), name (string), pattern (string), agg,
and flags. Return exactly three candidates for each assigned fid.
\end{verbatim}
\endgroup

\subsection{Alignment-based refinement}
\label{app:regex-refinement-details}

This prompt is used for the refinement stage in Section~\ref{sec:method-discover}, revising previously proposed regex feature extractors using fit-set feedback on their alignment with the corresponding latent family's activations.

\begingroup
\small
\begin{verbatim}
You are authoring regex feature extractors for latent families of a
sparse autoencoder. Work only from the supplied family briefs and
fit feedback. All family evidence comes from the unlabeled corpus.
No routing benchmark information is available. Do not use tools.
An external controller performs scoring and retention; return only
the candidates requested for this round.

INPUT
Batch {batch_id} of six; library width {N}; round 2.
<family_brief>
{fid, number of latents, activation share,
  Neuronpedia descriptions, highly associated tokens,
  family-specific tokens, punctuation/symbol tokens,
  digit-bearing tokens, for each assigned family}
</family_brief>
<round_request>
{"candidates_per_active_fid": {each assigned fid: 3},
 "previous_fit_scores": [
   {fid, name, pattern, agg, flags, fit, density, ceiling_fit}
   for each previous candidate]}
</round_request>
fit is Kendall tau-b on fit documents; density is match density.
ceiling_fit is the family's token-set reference score.

TASK
Propose three revised or alternative extractors per family whose
values better track the ranking of its member-latent firing-event
count across documents.
The controller scores candidates by Kendall tau-b on fit documents.
If brief tokens are content words, capture their shared form,
not the words themselves. Do not mechanically negate
anti-correlated values.

CONSTRAINTS
Use symbols, orthography, and structure: punctuation, digits,
operators, brackets, casing, whitespace/layout, Unicode classes,
run lengths, and closed-class code/LaTeX/HTML/SQL keywords.
Do not use English content-word literals, topical vocabularies,
or morphological suffix patterns. Use non-capturing groups (?:...).

AGGREGATORS
count: number of matches; ratio: matches per approximate word;
binary: whether any match exists; maxrun: longest match length.
Flags: "", "i", "m", or "im".

CONTROLLER PROCEDURE
After the initial three candidates, the controller requests three
refinement candidates together in one response, giving six proposed
candidates per family. Duplicates consume slots; there is no early
stopping or additional round. It retains the fit-best of all six,
preferring count/ratio within 0.01 of maximum fit.
Prefer ratio when fit scores are within 0.01 of the best.

OUTPUT
Return only a JSON array, without prose or markdown. Each object
must contain fid (integer), name (string), pattern (string), agg,
and flags. Return exactly three NEW candidates for each assigned
fid. Do not repeat the previous candidate array.
\end{verbatim}
\endgroup

\subsection{Authoring without SAE guidance}
\label{app:regex-control-template}

This prompt is used for the LLM-only control in Section~\ref{sec:ablation-authoring}, authoring regex feature extractors without family descriptions, associated tokens, or activation feedback.
The LLM proposes three candidates per slot in each of two rounds and selects its preferred candidate from all six, without SAE-based selection.

\begingroup
\small
\begin{verbatim}
You are authoring regex feature extractors. Work only from the
supplied slot list and previous candidates. No routing benchmark
information is available.
Do not use tools. An external controller performs retention;
return only the candidates requested.

INPUT
Batch {batch_id} of six; library width {N}; round {round_id}.
<slots>
{one object containing only fid for each assigned slot}
</slots>
<round_request>
{"candidates_per_fid": {each assigned fid: 3},
 "previous_candidates": [
   {fid, name, pattern, agg, flags} for each previous candidate]}
</round_request>
previous_candidates is empty in round 1 and contains the first
three candidates per slot in round 2. No evaluation feedback is given.

TASK
Propose three diverse extractors per slot measuring a surface-form
property of text. Different slots should measure different properties.
In round 2, propose three additional candidates and identify your
preferred candidate among all six for each slot. Number candidates
1-3 in the order previously returned and 4-6 in the order of the
new response. Base your choice on the instructions alone; no SAE
alignment or routing scores are available. The controller retains
your preferred candidate for each slot.

CONSTRAINTS
Use symbols, orthography, and structure: punctuation, digits,
operators, brackets, casing, whitespace/layout, Unicode classes,
run lengths, and closed-class code/LaTeX/HTML/SQL keywords.
Do not use English content-word literals, topical vocabularies,
or morphological suffix patterns. Use non-capturing groups (?:...).

AGGREGATORS
count: number of matches; ratio: matches per approximate word;
binary: whether any match exists; maxrun: longest match length.
Flags: "", "i", "m", or "im".

Prefer ratio when candidates are otherwise comparable, because
dense/rate-like features are wanted. Prefer count/ratio over
binary/maxrun when otherwise comparable.

OUTPUT
Return only a JSON object, without prose or markdown:
{"candidates": [...], "preferred_by_fid": {...}}.
The candidates array must contain exactly three NEW candidates
per assigned fid, each with fid (integer), name (string), pattern
(string), agg, and flags. In round 1, preferred_by_fid is empty.
In round 2, it maps each fid to your preferred candidate's index
from 1 to 6. Do not repeat the previous candidate array.
\end{verbatim}
\endgroup

\section{Experimental details}
\label{app:setup}
\begin{table}[!htbp]
\centering
\small
\caption{Candidate-pool size and the number of queries in each split.}
\label{tab:benchmarks}
\begin{tabular}{lrrrr}
\toprule
benchmark & \#models & train & validation & test \\
\midrule
EmbedLLM  & 112 & 29{,}673 & 3{,}000 & 3{,}000 \\
NineBy30k & 9   & 23{,}027 & 3{,}000 & 5{,}000 \\
R2Bench   & 10  & 18{,}580 & 3{,}097 & 9{,}291 \\
CARROT    & 13  & 30{,}968 & 6{,}636 & 6{,}637 \\
\bottomrule
\end{tabular}
\end{table}

\subsection{Benchmarks template removal}
\label{app:preprocessing}

\paragraph{Prompt templates.}
Every benchmark combines questions from several source datasets, and each source renders its questions inside a fixed prompt template.
In EmbedLLM, every MMLU question starts with ``The following are multiple choice questions (with answers) about $\langle$subject$\rangle$.'' and ends with ``Answer:''.
In NineBy30k, most multiple-choice questions start with ``Please read the following multiple-choice questions \dots'' and all end with the same \texttt{\textbackslash boxed\{X\}} instruction.
The template is the same for every query from a source and different across sources, so it identifies the source.
A router may therefore use the template as a shortcut, associating each source with a model that performs well on its training queries.
Such a router would also fail on real traffic, which does not come wrapped in benchmark prompts.
We therefore remove the templates from every benchmark before any representation sees the text, for our features and all baselines alike.
We modify only the query text and keep the original response scores.

\paragraph{Removal procedure.}
EmbedLLM records the source of each query, so we handle it per source. We remove the instruction header, keep only the final question for the few-shot sources (GSM8K, TruthfulQA), drop the \texttt{Q:}/\texttt{A:} frame of Social~IQa and the \texttt{Question:}, \texttt{Passage:}, and \texttt{Choices:} labels, remove the trailing \texttt{Answer:} cue, and unify the option markers.
Layout is a fingerprint too: multiple-choice sources put one option per line, generative sources one paragraph.
We therefore render every EmbedLLM query on one line with options marked \texttt{(A)}, \texttt{(B)}, \dots
The other three benchmarks have no source column, but their templates are leading instruction paragraphs and trailing format instructions, so one generic pass handles them: it removes leading paragraphs that start with an instruction (``Please answer \dots'', ``You are a knowledge expert \dots''), drops field labels (\texttt{Context:}, \texttt{Question:}, \texttt{Passage:}, \texttt{Title:}), and removes trailing answer cues and format instructions such as the \texttt{\textbackslash boxed\{X\}} block.
Queries without a wrapper pass through unchanged; almost every NineBy30k query has one, and about a third of R2Bench and CARROT queries do.
The question, any passage, and the answer options are kept verbatim, and the same cleaning is applied to all splits and to the perturbed test queries in Section~\ref{sec:robustness}.
For example, ``Please read the following multiple-choice questions and provide the most likely correct answer based on the options given. Context: None. Question: The value of log2 4 is: Options: A. 2 \dots\ E. 12. Provide the correct letter choice in \textbackslash boxed\{X\}, where X is the correct letter choice. Keep the explanation or feedback within 3 sentences.'' becomes ``The value of log2 4 is: Options: A. 2 \dots\ E. 12''.

\subsection{Pipeline configuration}
\label{app:pipeline-config}

\paragraph{Unlabeled discovery corpus.}
The discovery corpus $\mathcal{D}_{\mathrm{text}}$ contains 20,035 public documents covering web prose, mathematics, code, and instruction/dialogue text.
It comprises 5,000 documents each from FineWeb~\citep{penedo2024fineweb}, OpenWebMath~\citep{paster2023openwebmath}, CodeParrot, and OpenAssistant~\citep{kopf2023openassistant}, plus 35 instructions from Alpaca~\citep{taori2023alpaca}.
We use FineWeb's \texttt{sample-10BT} subset and the \texttt{codeparrot-clean-valid} release of CodeParrot.\footnote{\url{https://huggingface.co/datasets/codeparrot/codeparrot-clean-valid}}
Using random seed 0, we fix a split of 13,394 fit documents for regex refinement and selection and 6,641 held-out documents for alignment evaluation after the extractors are frozen.
We truncate each document to 512 tokens and evaluate candidate regexes on the corresponding decoded text span.
The same documents and fit/held-out split are used across discovery models.

\paragraph{Pretrained LMs and SAEs.}
We obtain pretrained SAE checkpoints from Gemma Scope~\citep{lieberum2024gemmascopeopensparse}\footnote{\url{https://huggingface.co/google/gemma-scope-2b-pt-res/tree/main/layer_25/width_16k/average_l0_116}}, Qwen-Scope~\citep{deng2026qwenscopeturningsparsefeatures}\footnote{\url{https://huggingface.co/Qwen/SAE-Res-Qwen3-1.7B-Base-W32K-L0_100/blob/main/layer27.sae.pt}}, and Llama Scope~\citep{he2024llamascopeextractingmillions}\footnote{\url{https://huggingface.co/OpenMOSS-Team/Llama3_1-8B-Base-LXR-8x/tree/main/Llama3_1-8B-Base-L31R-8x}} for Gemma-2-2B, Qwen3-1.7B-Base, and Llama-3.1-8B, respectively.
The LMs and SAEs remain frozen, and each SAE encodes the residual stream after the final transformer block.
Their activations supply the latent families and alignment targets used in regex discovery.
The default configuration uses Gemma-2-2B with its 16,384-latent SAE.
Section~\ref{sec:ablation-source} evaluates alternative discovery LM--SAE pairs, with per-benchmark results in Appendix~\ref{app:discovery-scaling-results}.

\FloatBarrier
\subsection{Query representations}
\label{app:encoders}

Table~\ref{tab:encoders} lists the neural encoders by category, with their checkpoints, dimensions, pooling methods, and supported context lengths.
TF-IDF uses character 3--5-grams (\texttt{char\_wb}, minimum document frequency 3, up to 40{,}000 terms) and word 1--2-grams (minimum document frequency 2, up to 20{,}000 terms) with sublinear term frequency and $\ell_2$ normalization.
For the scaling study, we encode every query with the Qwen2.5 base models in one pipeline: a prefill-only forward pass in bf16, last-token pooling, and an 8{,}192-token context with head truncation.

\begin{table}[!htbp]
\centering
\footnotesize
\caption{\textbf{Learned query representations.} All checkpoints are frozen.}
\label{tab:encoders}
\setlength{\tabcolsep}{4pt}
\resizebox{\linewidth}{!}{%
\begin{tabular}{lllrll}
\toprule
category & representation & checkpoint & dim & pooling & context length \\
\midrule
\multirow{5}{*}{\shortstack[l]{Sentence-embedding\\models}}
 & MiniLM-L6        & \texttt{all-MiniLM-L6-v2}  & 384  & mean & 512 \\
 & MPNet-base       & \texttt{all-mpnet-base-v2} & 768  & mean & 512 \\
 & BGE-base         & \texttt{BAAI/bge-base-en-v1.5}                    & 768  & \texttt{[CLS]} & 512 \\
 & GTE-base         & \texttt{thenlper/gte-base}                        & 768  & mean & 512 \\
 & E5-base          & \texttt{intfloat/e5-base-v2}                      & 768  & mean & 512 \\
\midrule
\multirow{2}{*}{Encoder-only LMs}
 & ModernBERT-base  & \texttt{answerdotai/ModernBERT-base}              & 768  & \texttt{[CLS]} & 8{,}192 \\
 & ModernBERT-large & \texttt{answerdotai/ModernBERT-large}             & 1024 & \texttt{[CLS]} & 8{,}192 \\
\midrule
\multirow{2}{*}{Decoder-only LMs}
 & Qwen2.5-0.5B     & \texttt{Qwen/Qwen2.5-0.5B}                        & 896  & last token & 32{,}768 \\
 & Qwen2.5 ladder   & \texttt{Qwen/Qwen2.5-\{0.5B..72B\}} & 896--8192 & last token & 32{,}768--131{,}072 \\
\bottomrule
\end{tabular}}
\end{table}

\FloatBarrier
\subsection{Encoder scaling setup}
\label{app:scaling}
\label{sec:encoder-scaling}

We compare frozen Qwen2.5 base models at seven sizes, from 0.5B to 72B parameters, using the uniform encoding pipeline described in Appendix~\ref{app:encoders}.
For each model and benchmark, we train an Extra-Trees routing head and select its hyperparameters on validation data from the search space in Appendix~\ref{app:heads}, using the same six configurations at every model size.
We repeat each fit with five random seeds and report the mean test accuracy.
The other neural encoders and the 128 regex features in Figure~\ref{fig:encoder-scaling}b follow the same protocol.
Figure~\ref{fig:encoder-scaling}b reports the unweighted mean test routing accuracy across the four benchmarks.
Despite a 147$\times$ increase in model size, it ranges only from 76.31\% to 76.52\%, and the 128 regex features reach 76.41\% without running a neural text encoder at inference.

\FloatBarrier
\subsection{Robustness details}
\label{app:robustness}

\paragraph{Perturbations.}
We generate one rewrite per test query and mode with DeepSeek-V4-Flash through OpenRouter, following the RouterArena~\citep{lu2025routerarena}.
The system prompt requires every fact and answer option to be preserved.
Misspellings introduce spelling errors while keeping the text readable; word substitutions replace content words with synonyms; sentence restructuring changes sentence structure; and paraphrasing rewrites the query using different wording.
Perturbed queries are assigned the original response scores.
Table~\ref{tab:robust} reports routing accuracy under each rewrite.

\begin{table}[!htb]
\centering
\small
\caption{\textbf{Routing accuracy under LLM query perturbation (\%, four-benchmark mean, Extra-Trees).} Fitted on clean text; only test queries are perturbed. Clean reference scores are from Table~\ref{tab:main}; \emph{drop} is the perturbation mean minus this reference. Accuracy-column maxima in \textbf{bold}.}
\label{tab:robust}
\begin{tabular}{lcccccr}
\toprule
representation & clean & typo & synonym & grammar & paraphrase & drop \\
\midrule
128 regexes (ours) & \textbf{76.43} & \textbf{76.43} & \textbf{75.74} & 75.41 & 74.92 & $-0.81$ \\
TF-IDF bags        & 75.70 & 75.27 & 74.43 & 75.06 & 74.27 & $-0.94$ \\
Qwen2.5-0.5B       & 76.33 & 75.80 & 75.28 & \textbf{75.71} & \textbf{75.18} & $-0.84$ \\
ModernBERT-large   & 75.91 & 74.84 & 74.76 & 75.12 & 74.81 & $-1.03$ \\
ModernBERT-base    & 75.58 & 75.07 & 74.70 & 75.06 & 74.44 & $-0.76$ \\
GTE-base           & 75.09 & 74.58 & 74.68 & 74.91 & 74.72 & $-0.37$ \\
BGE-base           & 75.07 & 74.69 & 74.48 & 74.76 & 74.45 & $-0.47$ \\
MPNet-base         & 74.96 & 74.81 & 74.69 & 74.92 & 74.63 & $-0.20$ \\
MiniLM-L6          & 75.02 & 74.64 & 74.47 & 74.71 & 74.48 & $-0.45$ \\
E5-base            & 74.92 & 74.50 & 74.40 & 74.63 & 74.50 & $-0.41$ \\
\bottomrule
\end{tabular}
\end{table}

\begin{figure}[!htb]
\centering
\includegraphics[width=0.62\linewidth]{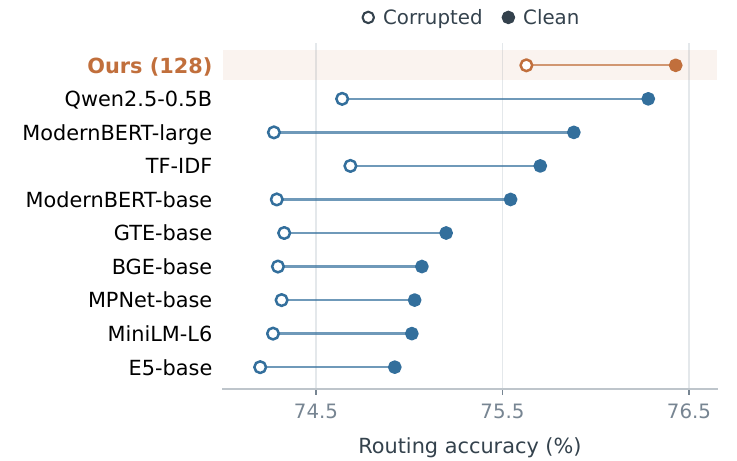}
\caption{\textbf{Routing accuracy under mechanical character corruption} (\%, four-benchmark mean, Extra-Trees head), using the marker convention of Figure~\ref{fig:robust-latency} (left): filled is clean, hollow is the mean over corruption rates of 5, 10, 20, and 30\% of characters.}
\label{fig:robust-corruption}
\end{figure}

\paragraph{Character corruption.}
We generate corrupted test queries with a Python script. At each alphabetic character, the script applies an edit with probability $p\in\{0.05,0.10,0.20,0.30\}$, choosing uniformly among substitution with a neighboring keyboard character, deletion, swapping with the next character, and duplication. We fix the random seed to 0 and evaluate the routers trained on clean text using the original response scores.
Figure~\ref{fig:robust-corruption} and Table~\ref{tab:typo-sweep} report results across corruption levels.
Relative to the clean reference scores in Table~\ref{tab:main}, accuracy at 30\% corruption is lower by 1.43 percentage points for the regex router and 0.96--1.16 percentage points for the sentence-embedding models.
Nevertheless, it achieves the highest routing accuracy among the evaluated representations at every corruption level.
Qwen2.5-0.5B shows the largest decrease, at 2.28 percentage points.

\begin{table}[!htb]
\centering
\small
\caption{\textbf{Mechanical character corruption (\%, four-benchmark mean, Extra-Trees).} Clean reference scores are from Table~\ref{tab:main}. Ours and four representative baselines; the remaining encoders are in Figure~\ref{fig:robust-corruption}.}
\label{tab:typo-sweep}
\begin{tabular}{lccccc}
\toprule
representation & clean & 5\% & 10\% & 20\% & 30\% \\
\midrule
128 regexes (ours) & 76.43 & 76.13 & 75.97 & 75.41 & 75.00 \\
TF-IDF bags        & 75.70 & 75.00 & 74.96 & 74.41 & 74.36 \\
Qwen2.5-0.5B       & 76.33 & 75.30 & 74.88 & 74.33 & 74.05 \\
E5-base            & 74.92 & 74.45 & 74.44 & 74.06 & 73.85 \\
MiniLM-L6          & 75.02 & 74.63 & 74.43 & 74.16 & 73.86 \\
\bottomrule
\end{tabular}
\end{table}

\FloatBarrier
\subsection{Latency details}
\label{app:latency}

We measure latency on test queries with a batch size of one, using a machine with an AMD EPYC 7513 CPU and eight NVIDIA RTX A5000 GPU. CPU measurements use eight threads pinned to CPU cores.
At $N=128$, one query costs 3.14\,ms of regex extraction, 1.05\,ms of forest prediction, and 0.01\,ms of standardization. Regex feature extraction grows linearly with $N$.

\begin{table}[!htb]
\centering
\small
\caption{\textbf{Routing latency and GPU memory use.} CARROT queries, batch size one. Timings include feature extraction and routing-head inference.}
\label{tab:latency}
\begin{tabular}{@{}lrrr@{}}
\toprule
Router & p50 (ms) & p99 (ms) & VRAM (MB) \\
\midrule
\multicolumn{4}{@{}l}{\textbf{GPU}} \\
\addlinespace[2pt]
MiniLM-L6          & 5.0 & 10.4 & 204 \\
E5-base            & 8.2 & 13.6 & 832 \\
Qwen2.5-0.5B       & 19.5 & 25.8 & 2{,}617 \\
ModernBERT-large   & 21.6 & 27.7 & 1{,}609 \\
\midrule
\multicolumn{4}{@{}l}{\textbf{CPU}} \\
\addlinespace[2pt]
MiniLM-L6          & 11.6 & 40.6 & --- \\
E5-base            & 53.4 & 256.6 & --- \\
Qwen2.5-0.5B       & 200.5 & 846.0 & --- \\
ModernBERT-large   & 219.6 & 863.8 & --- \\
\addlinespace[2pt]
Ours ($N=32$)      & 1.4 & 11.9 & --- \\
Ours ($N=128$)     & 2.2 & 37.1 & --- \\
\bottomrule
\end{tabular}
\end{table}

The combined test queries from our four benchmarks have a median length of 249 characters and a 99th-percentile length of 6{,}607 characters.
Table~\ref{tab:latency-traffic} reports latency on a sample of 2{,}979 queries from these benchmarks, reweighted to match this length distribution.
The p99 of the 128-feature router exceeds the GPU baselines, while the 32-feature router stays below all of them except MiniLM-L6 and E5-base.

\begin{table}[!htb]
\centering
\small
\caption{\textbf{Traffic-weighted latency (ms)} on a pooled trace of 2{,}979 queries, batch size one, reweighted to the real query-length distribution.}
\label{tab:latency-traffic}
\begin{tabular}{llrrr}
\toprule
router & device & p50 & p90 & p99 \\
\midrule
32 regexes (ours)  & CPU & 1.8 & 3.8 & 11.5 \\
128 regexes (ours) & CPU & 3.0 & 8.6 & 33.6 \\
MiniLM-L6          & GPU & 4.9 & 5.4 & 7.8 \\
E5-base            & GPU & 8.2 & 8.9 & 10.8 \\
Qwen2.5-0.5B       & GPU & 19.4 & 20.1 & 22.6 \\
MiniLM-L6          & CPU & 13.2 & 30.7 & 36.3 \\
E5-base            & CPU & 57.8 & 147.3 & 198.6 \\
\bottomrule
\end{tabular}
\end{table}

\FloatBarrier
\section{Ablation Details}
\label{app:ablations}

\subsection{Choice of discovery LM and SAE}
\label{app:discovery-scaling-results}

Table~\ref{tab:source} gives the per-benchmark scores and SAE dictionary sizes for the discovery-model comparison in Figure~\ref{fig:ablation-overview}(a).
On each benchmark, routing accuracy is similar across the evaluated discovery LM--SAE pairs.
The pretrained SAE checkpoints and discovery corpus are described in Appendix~\ref{app:pipeline-config}.

\begin{table}[!htbp]
\centering
\small
\setlength{\tabcolsep}{4pt}
\caption{\textbf{Discovery LM and SAE.} Test routing accuracy (\%) with $N=128$. Each row corresponds to a discovery LM--SAE pair, and $K$ is the number of latents in the SAE.}
\label{tab:source}
\begin{tabular}{lrccccc}
\toprule
Source LM & $K$ & EmbedLLM & NineBy30k & R2Bench & CARROT & Avg. \\
\midrule
Gemma-2-2B & 16{,}384 & 67.37 & 69.01 & 82.16 & 87.18 & 76.43 \\
Qwen3-1.7B & 32{,}768 & 67.53 & 68.91 & 82.15 & 87.18 & 76.44 \\
Llama-3.1-8B & 32{,}768 & 67.07 & 69.27 & 81.95 & 87.19 & 76.37 \\
\bottomrule
\end{tabular}
\end{table}

\FloatBarrier
\subsection{Choice of latent grouping}
\label{app:grouping}

We compare spectral clustering with three alternatives: spherical $k$-means, facility-location selection, and average-linkage agglomerative clustering.
All four methods group the same 16{,}384 Gemma SAE latents using their document-level activation vectors on the 13{,}394 fit documents.
For each method and $N$, we construct a regex library using the same authoring budget and alignment feedback, then evaluate it with an Extra-Trees routing head.

Spectral clustering retains the 50 largest absolute correlations per latent, symmetrizes the graph, and clusters the resulting spectral embedding into $N$ families.
Spherical $k$-means clusters mean-centered, unit-normalized activation vectors using $k$-means++ initialization, 60 iterations, and random seed 0.
Agglomerative clustering uses average linkage with dissimilarity equal to one minus the absolute correlation.
Facility-location selection greedily chooses $N$ representative latents to maximize coverage, weighted by each latent's activation variance across documents.
Coverage is measured by a latent's largest absolute correlation with any selected representative.
Each latent is then assigned to the representative with the highest absolute correlation.

Table~\ref{tab:grouping-curves} reports mean test routing accuracy across four benchmarks for each method and feature count $N$.
Table~\ref{tab:grouping-family-sizes} reports routing accuracy and family-size statistics at $N=128$, where spectral clustering produces the smallest maximum family.
We prefer a small maximum family size because each family is approximated by a single extractor, and a very large family combines many distinct properties that one surface pattern is unlikely to track.

\begin{table}[!htbp]
\centering\small
\caption{Mean test routing accuracy (\%) across four benchmarks by grouping method and feature count $N$. Each cell reports a separately authored regex library evaluated with an Extra-Trees head.}
\label{tab:grouping-curves}
\begin{tabular}{@{}rrrrr@{}}
\toprule
$N$ & Spectral & Spherical $k$-means & Agglomerative & Facility location \\
\midrule
1 & 73.48 & 73.64 & \textbf{73.66} & 73.65 \\
2 & 73.80 & 73.82 & 73.80 & \textbf{73.96} \\
4 & \textbf{75.08} & 74.41 & 73.73 & 73.89 \\
8 & \textbf{75.24} & 74.54 & 74.95 & 74.95 \\
16 & 75.62 & 74.80 & 75.35 & \textbf{75.66} \\
32 & 76.06 & 75.23 & \textbf{76.12} & 75.69 \\
64 & 75.98 & \textbf{76.25} & 76.03 & 75.97 \\
128 & \textbf{76.43} & 76.10 & 76.03 & 76.03 \\
256 & 76.19 & \textbf{76.39} & 76.25 & 76.25 \\
512 & \textbf{76.31} & 76.15 & 76.31 & 76.14 \\
1,024 & 76.31 & 76.17 & 76.28 & \textbf{76.32} \\
\bottomrule
\end{tabular}
\end{table}

\begin{table}[!htbp]
\centering
\small
\caption{Routing performance and family sizes at $N=128$ with different clustering strategies.}
\label{tab:grouping-family-sizes}
\begin{tabular}{@{}lrrrr@{}}
\toprule
 & & \multicolumn{3}{c}{Latents per family} \\
\cmidrule(lr){3-5}
Grouping method & Accuracy (\%) & Minimum & Median & Maximum \\
\midrule
Spectral clustering & 76.43 & 31 & 115.0 & 439 \\
Spherical $k$-means & 76.10 & 22 & 95.5 & 531 \\
Agglomerative clustering & 76.03 & 1 & 15.0 & 4{,}904 \\
Facility-location selection & 76.03 & 1 & 56.5 & 1{,}831 \\
\bottomrule
\end{tabular}
\end{table}

\FloatBarrier
\subsection{SAE guidance and refinement}
\label{app:authoring-ablations}

The ablation in Section~\ref{sec:ablation-authoring} gives every variant the same budget of six candidate extractors per output feature.
We keep the authoring model, regex interface, vocabulary constraints, and routing evaluation protocol fixed across variants.
We use the corresponding prompt templates in Appendix~\ref{app:regex-authoring-prompt}.
Each variant proposes three candidates together in each of two rounds and retains one of the six, yielding a library of $N$ features.
Duplicate proposals consume candidate slots, and no variant stops early or receives replacement proposals.

The LLM-only variant receives no SAE evidence or alignment feedback and retains the LLM's preferred candidate among all six for each feature.
The two SAE-guided variants use the same Gemma-2-2B SAE partition and family briefs.
Here, refinement refers to returning alignment feedback to the authoring LLM so it can revise its proposals.
The variant without refinement (\emph{initial regexes} in Table~\ref{tab:authoring-ablation}) proposes all six candidates without receiving fit feedback.
The full procedure instead returns all three initial candidates and their fit scores to the LLM before it proposes three revised or alternative candidates.
Both guided variants select one extractor from all six using Kendall $\tau_b$ on the unlabeled fit corpus, with the same aggregation preferences specified in Appendix~\ref{app:regex-authoring-prompt}.

Comparing LLM-only with the variant without refinement measures the combined effect of SAE-family information and alignment-based candidate selection.
Comparing the two SAE-guided variants measures the additional effect of feedback-driven revision, with the candidate budget and final selection rule held fixed.

To compare alignment across libraries, we select a one-to-one regex--family matching that maximises mean Kendall $\tau_b$ on fit documents, then report its mean alignment on held-out documents.
This matching is used only for evaluation after authoring.
It does not provide SAE feedback to the LLM-only control or change the extractors or routing results.

\FloatBarrier
\subsection{Routing-head details}
\label{app:heads}

We use an Extra-Trees regressor with at least 25 samples per leaf, trained on the full per-model score vector.
For each representation and benchmark, we select the number of trees, maximum depth, and features per split on validation from $\{(300,\infty,1.0),(300,12,1.0),(300,24,1.0),(100,\infty,1.0),(300,\infty,\sqrt{d}),(100,\infty,\sqrt{d})\}$, where $1.0$ means all $d$ coordinates are considered at every split, and read test once.
For the ablation in Section~\ref{sec:ablation-heads}, we additionally evaluate a decision tree (maximum leaves in $\{8,16,32,64,128\}$), a random forest using the forest grid above, cosine $k$-nearest neighbors ($k\in\{10,25,50,100\}$), and mini-batch $k$-means ($K\in\{50,100,150,200\}$).

\begin{table}[!htbp]
\centering
\small
\caption{\textbf{Choice of routing head.} Mean test routing accuracy (\%) across four benchmarks, with hyperparameters selected on validation. RF: random forest; ET: Extra-Trees. Bold marks the best representation for each head.}
\label{tab:heads}
\begin{tabular}{lccccc}
\toprule
Representation & Tree & RF & ET & $k$-NN & $k$-means \\
\midrule
128 regexes (ours) & \textbf{75.17} & \textbf{76.37} & \textbf{76.43} & 75.01 & 75.09 \\
TF-IDF bags        & 74.41 & 75.57 & 75.70 & 74.73 & 74.68 \\
Llama-3.1-8B       & 74.37 & 76.18 & 76.41 & 75.77 & 75.61 \\
Gemma-2-2B         & 74.31 & 76.11 & 76.39 & 75.85 & 75.58 \\
Qwen2.5-0.5B       & 74.54 & 76.10 & 76.33 & \textbf{76.04} & \textbf{75.73} \\
ModernBERT-large   & 74.08 & 75.68 & 75.91 & 75.40 & 75.41 \\
ModernBERT-base    & 74.29 & 75.46 & 75.58 & 75.40 & 75.28 \\
GTE-base           & 73.66 & 75.05 & 75.09 & 75.16 & 74.94 \\
BGE-base           & 73.42 & 74.87 & 75.07 & 75.31 & 75.02 \\
MiniLM-L6          & 73.49 & 74.81 & 75.02 & 75.28 & 74.83 \\
MPNet-base         & 73.92 & 74.81 & 74.96 & 75.12 & 74.93 \\
E5-base            & 73.60 & 74.78 & 74.92 & 75.57 & 75.09 \\
\bottomrule
\end{tabular}
\end{table}

\section{Complete Regex Feature List}
\label{app:regex-features}

Table~\ref{tab:regex-features} enumerates all 128 frozen regex features.
The indices f1--f128 follow the category order used in Figure~\ref{fig:source-feature-centers}.
The eight categories describe surface properties rather than task types and are distinct from the SAE latent families used during discovery.
These regex feature extractors are obtained from Gemma-2-2B using the pipeline described in Section~\ref{sec:method}.

For a query $q$, let $c_j(q)$ be the number of non-overlapping matches returned by Python's \texttt{re.findall} for pattern $p_j$.
The library uses two aggregation rules:
\begin{equation}
    \phi_j(q)=
    \begin{cases}
        c_j(q), & a_j=\texttt{count},\\
        c_j(q)/\max\!\left(|q|/5,1\right), & a_j=\texttt{ratio},
    \end{cases}
    \label{eq:appendix-regex-aggregation}
\end{equation}
where $|q|$ is the number of characters in the query.
Thus, \texttt{ratio} is a count per approximate word, using five characters per word.
There are 124 \texttt{count} features and 4 \texttt{ratio} features.
The additional flags \texttt{i} and \texttt{m} denote case-insensitive and multiline matching, respectively. 
A dash means no additional flags.
For portable typesetting, literal non-ASCII characters are written as equivalent Python regex Unicode escapes (\texttt{\textbackslash uXXXX}).

\begingroup
\small
\setlength{\tabcolsep}{4pt}
\renewcommand{\arraystretch}{1.05}
\begin{longtable}{@{}p{26pt}p{\dimexpr\linewidth-26pt-2\tabcolsep\relax}@{}}
\caption{Complete feature library in the display order of Figure~\ref{fig:source-feature-centers}. Each entry gives its name, aggregation rule, additional matching flags, and full regex.}
\label{tab:regex-features}\\
\toprule
\textbf{Index} & \textbf{Feature name}\hfill\textbf{Aggregation; flags}\\
\midrule
\endfirsthead
\multicolumn{2}{l}{\small\textit{Table~\thetable\ continued}}\\
\toprule
\textbf{Index} & \textbf{Feature name}\hfill\textbf{Aggregation; flags}\\
\midrule
\endhead
\midrule
\multicolumn{2}{r}{\small\textit{Continued on next page}}\\
\endfoot
\bottomrule
\endlastfoot
\multicolumn{2}{@{}l}{\rule{0pt}{2.2ex}\textbf{Script (12 features)}}\\*
f1 & \texttt{cat\_accent\_elision\_union}\hfill\texttt{count}; ---\\*
 & {\footnotesize\ttfamily\raggedright (\allowbreak{}?\allowbreak{}:\allowbreak{}[\allowbreak{}\textbackslash{}u00e0\allowbreak{}\textbackslash{}u00e8\allowbreak{}\textbackslash{}u00e9\allowbreak{}\textbackslash{}u00ed\allowbreak{}\textbackslash{}u00f2\allowbreak{}\textbackslash{}u00f3\allowbreak{}\textbackslash{}u00fa\allowbreak{}\textbackslash{}u00ef\allowbreak{}\textbackslash{}u00e7\allowbreak{}\textbackslash{}u00c0\allowbreak{}\textbackslash{}u00c8\allowbreak{}\textbackslash{}u00c9\allowbreak{}\textbackslash{}u00cd\allowbreak{}\textbackslash{}u00d2\allowbreak{}\textbackslash{}u00d3\allowbreak{}\textbackslash{}u00da\allowbreak{}\textbackslash{}u00cf\allowbreak{}\textbackslash{}u00c7\allowbreak{}\textbackslash{}u00b7\allowbreak{}]\allowbreak{}|\allowbreak{}\textbackslash{}b\allowbreak{}[\allowbreak{}L\allowbreak{}l\allowbreak{}D\allowbreak{}d\allowbreak{}S\allowbreak{}s\allowbreak{}N\allowbreak{}n\allowbreak{}M\allowbreak{}m\allowbreak{}T\allowbreak{}t\allowbreak{}]\allowbreak{}[\allowbreak{}\textbackslash{}u2019\allowbreak{}'\allowbreak{}]\allowbreak{}(\allowbreak{}?\allowbreak{}=\allowbreak{}[\allowbreak{}a\allowbreak{}-\allowbreak{}z\allowbreak{}\textbackslash{}u00e0\allowbreak{}-\allowbreak{}\textbackslash{}u00ff\allowbreak{}]\allowbreak{})\allowbreak{})\par}\\[4pt]
f2 & \texttt{it\_accent\_ratio}\hfill\texttt{ratio}; ---\\*
 & {\footnotesize\ttfamily\raggedright [\allowbreak{}\textbackslash{}u00e0\allowbreak{}\textbackslash{}u00e8\allowbreak{}\textbackslash{}u00e9\allowbreak{}\textbackslash{}u00ec\allowbreak{}\textbackslash{}u00f2\allowbreak{}\textbackslash{}u00f9\allowbreak{}\textbackslash{}u00c0\allowbreak{}\textbackslash{}u00c8\allowbreak{}\textbackslash{}u00c9\allowbreak{}\textbackslash{}u00cc\allowbreak{}\textbackslash{}u00d2\allowbreak{}\textbackslash{}u00d9\allowbreak{}]\par}\\
\end{longtable}
\endgroup

\clearpage
\section{Source Dataset Feature Profiles}
\label{app:source-profiles}

Figures~\ref{fig:appendix-source-profiles-embedllm}--\ref{fig:appendix-source-profiles-carrot-r2bench} extend Figure~\ref{fig:source-feature-centers} with the ten largest source datasets in each query pool.
In EmbedLLM, PIQA peaks on short-line counts, whereas MMLU professional law has higher word and long-word counts.
In the shared CARROT/R2Bench pool, MATH is elevated on LaTeX and mathematical-symbol features, while MuSR sources have higher sentence-boundary and punctuation counts.
These examples show that datasets have distinct average profiles across our regex coordinates, providing interpretable cues for distinguishing their queries.

Each bar is a source's mean feature value after standardizing query values with the corresponding training statistics and clipping to $[-4,4]$.
All panels use the same order of surface properties as Figure~\ref{fig:source-feature-centers}.
The complete list of the extractors is in Appendix~\ref{app:regex-features}.

\medskip
\noindent\begin{minipage}{\linewidth}
  \centering
  \includegraphics[width=\linewidth]{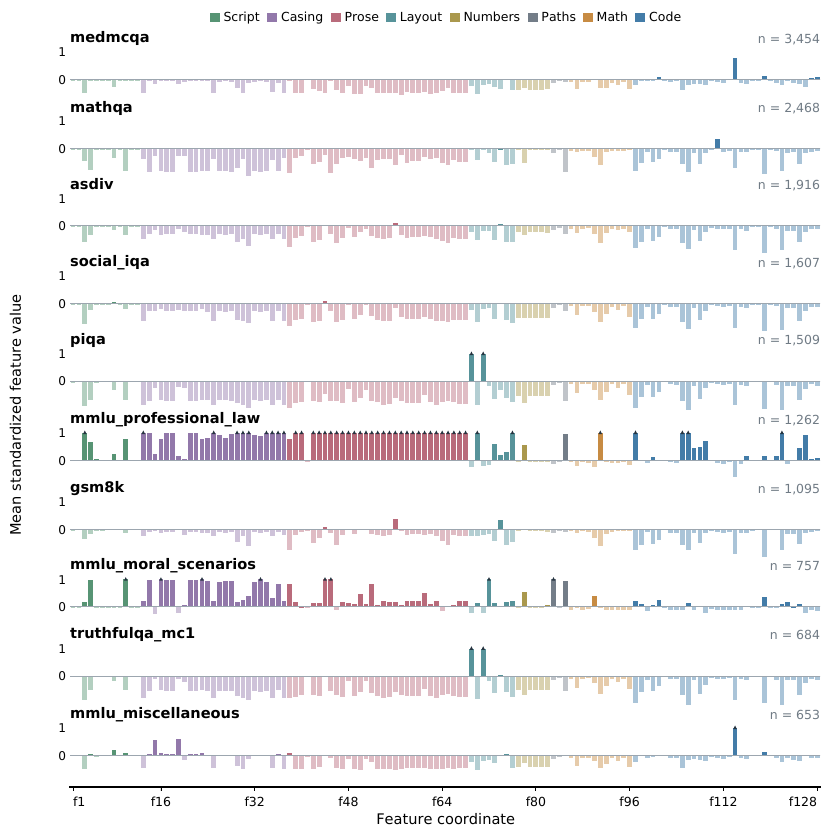}
  \captionof{figure}{\textbf{Source feature profiles for EmbedLLM.}
  The ten largest training sources, with one bar per feature. Triangles mark
  means above the $+1$ display limit; $n$ denotes the source's query count.}
  \label{fig:appendix-source-profiles-embedllm}
\end{minipage}

\clearpage
\noindent\begin{minipage}{\linewidth}
  \centering
  \includegraphics[width=\linewidth]{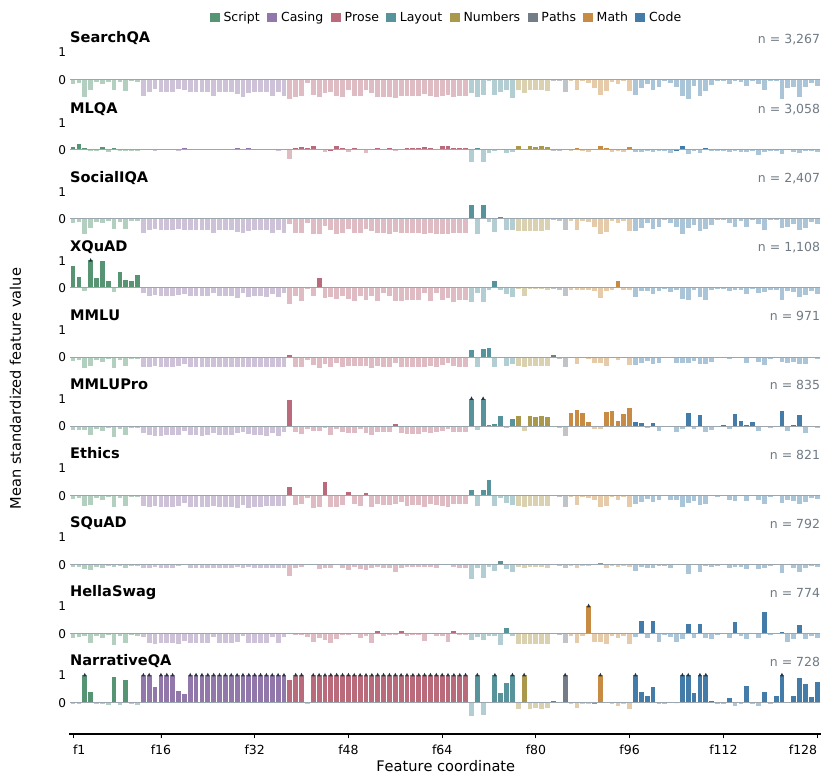}
  \captionof{figure}{\textbf{Source feature profiles for NineBy30k.}
  The ten largest training sources, with the same feature order, colors,
  and display limits as Figure~\ref{fig:appendix-source-profiles-embedllm}.}
  \label{fig:appendix-source-profiles-nineby30k}
\end{minipage}

\clearpage
\noindent\begin{minipage}{\linewidth}
  \centering
  \includegraphics[width=\linewidth]{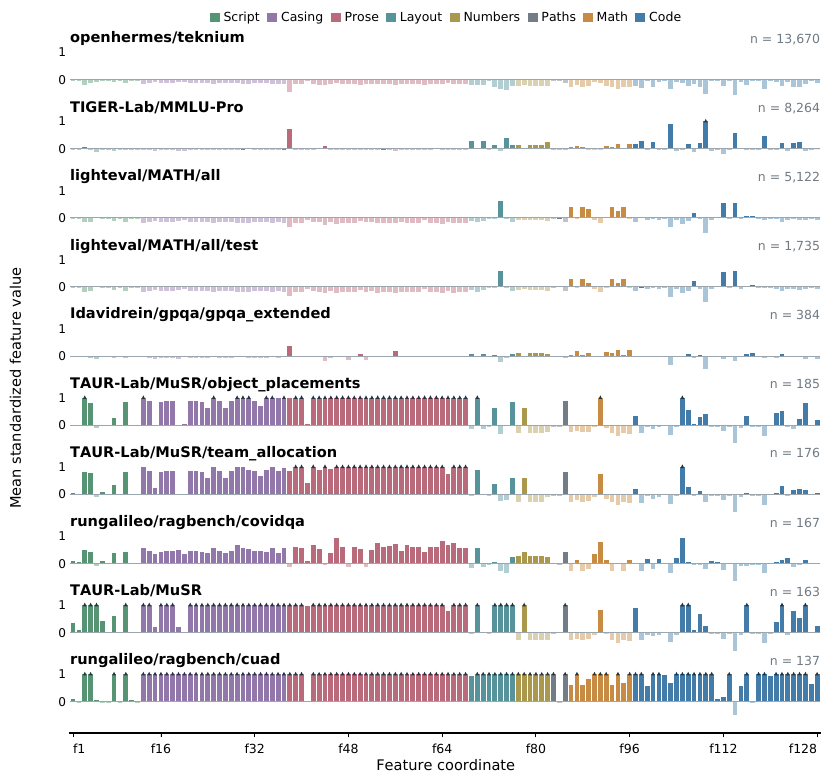}
  \captionof{figure}{\textbf{Shared source feature profiles for CARROT and R2Bench.}
  The ten largest source groups in CARROT training, which contains R2Bench's
  full query pool. Counts and scaling use this shared pool.}
  \label{fig:appendix-source-profiles-carrot-r2bench}
\end{minipage}
\clearpage

\section{Query Neighborhoods}
\label{app:query-neighborhoods}

\subsection{Setup and definitions}

We provide the setup and per-benchmark results for Figure~\ref{fig:query-neighborhoods}(a,b), comparing our 128 regex features with E5-base, ModernBERT-base, and Qwen2.5-0.5B.
We use the full test splits in Table~\ref{tab:benchmarks} and select $k=20$ neighbors for each query from the other test queries.

\paragraph{Forest neighborhoods.}
As in the main analysis, neighbors are defined by shared leaves in an Extra-Trees router.
For each representation, we fit the router to training model-score vectors, using 100 trees, unrestricted depth, square-root feature subsampling, minimum leaf size 25, squared-error loss, and seed zero.
These settings are fixed across representations for this analysis.
We pass test queries through the frozen forest and rank neighbors by the fraction of trees in which they share a terminal leaf with the query, comparing leaf identities within each tree.

\paragraph{Evaluation metrics.}
Let $d_i$ be query $i$'s original source label and $r_{i,m}\in[0,1]$ its observed response score for model $m$.
For a query pair $(i,j)$, we define source agreement and model-score similarity as
\begin{equation}
 A(i,j)=\mathbf{1}[d_i=d_j],
 \qquad
 S(i,j)=1-\frac{1}{M}\sum_{m=1}^{M}|r_{i,m}-r_{j,m}|.
\end{equation}
We average each metric over the same $k$ neighbors and all test queries, including those with constant score vectors, and report percentages.
The random reference averages each metric over all distinct test-query pairs within a benchmark.

\paragraph{Direct nearest neighbors.}
As an additional comparison, we select neighbors by Euclidean distance in each representation before fitting a router.
We standardize each coordinate using its training mean and standard deviation, with scale one for zero-variance coordinates.
The regex inputs are already z-scored and clipped to $[-4,4]$; we standardize these clipped inputs again before computing distances.
We use the same $k=20$, test queries, and evaluation metrics as for forest neighborhoods.

\subsection{Nearest-neighbor results}

Table~\ref{tab:query-neighborhoods-forest-full} reports the per-benchmark results underlying Figure~\ref{fig:query-neighborhoods}(a,b).
With forest neighborhoods, \ours{} reaches 68.24\% mean source agreement and 64.97\% model-score similarity, compared with 68.09\% and 65.48\% for Qwen2.5-0.5B.
Both means exceed those of E5-base and ModernBERT-base, as well as the random reference of 17.55\% and 57.74\%, respectively.

Tables~\ref{tab:query-neighborhoods-direct-source} and~\ref{tab:query-neighborhoods-direct-scores} report direct Euclidean neighbors.
Here, \ours{} reaches 67.00\% source agreement and 64.62\% model-score similarity, while Qwen2.5-0.5B reaches 72.27\% and 65.62\%.
Thus, regex features group queries by source and model performance under both definitions, although Qwen has stronger direct neighborhoods.

\begin{table}[!htbp]
  \centering
  \small
  \setlength{\tabcolsep}{4pt}
  \caption{\textbf{Forest neighborhoods, per benchmark.}
  Source agreement and model-score similarity (\%) at $k=20$
  for the representations in Figure~\ref{fig:query-neighborhoods}(a,b).}
  \label{tab:query-neighborhoods-forest-full}
  \textbf{A. Source agreement}\par\smallskip
\begin{tabular}{@{}lrrrrr@{}}
\toprule
Representation & EmbedLLM & NineBy30k & R2Bench & CARROT & Mean \\
\midrule
Random & 3.91 & 6.72 & 30.08 & 29.48 & 17.55 \\
E5-base & 24.71 & 35.77 & 72.46 & 66.29 & 49.81 \\
ModernBERT-base & 38.78 & 58.57 & 81.67 & 79.08 & 64.53 \\
Qwen2.5-0.5B & 44.82 & 60.12 & 84.84 & 82.59 & 68.09 \\
RegexRoute & 40.02 & 64.11 & 85.12 & 83.72 & 68.24 \\
\bottomrule
\end{tabular}
\par\medskip\textbf{B. Model-score similarity}\par\smallskip
\begin{tabular}{@{}lrrrrr@{}}
\toprule
Representation & EmbedLLM & NineBy30k & R2Bench & CARROT & Mean \\
\midrule
Random & 53.34 & 53.04 & 62.72 & 61.83 & 57.74 \\
E5-base & 58.60 & 57.14 & 69.44 & 68.47 & 63.41 \\
ModernBERT-base & 59.92 & 58.75 & 70.60 & 69.80 & 64.77 \\
Qwen2.5-0.5B & 60.46 & 59.40 & 71.45 & 70.62 & 65.48 \\
RegexRoute & 59.79 & 59.07 & 70.66 & 70.35 & 64.97 \\
\bottomrule
\end{tabular}

\end{table}

\begin{table}[!htbp]
  \centering
  \small
  \setlength{\tabcolsep}{4pt}
  \caption{\textbf{Source agreement under direct Euclidean neighborhoods.}
  Percentage of $k=20$ other test-query neighbors sharing the anchor's original
  dataset label, after standardizing coordinates using training statistics.}
  \label{tab:query-neighborhoods-direct-source}
  \begin{tabular}{@{}lrrrrr@{}}
\toprule
Representation & EmbedLLM & NineBy30k & R2Bench & CARROT & Mean \\
\midrule
Random & 3.91 & 6.72 & 30.08 & 29.48 & 17.55 \\
E5-base & 44.75 & 58.33 & 83.94 & 82.56 & 67.40 \\
ModernBERT-base & 41.88 & 64.73 & 83.99 & 82.53 & 68.28 \\
Qwen2.5-0.5B & 49.46 & 69.51 & 85.60 & 84.53 & 72.27 \\
RegexRoute & 37.80 & 63.05 & 84.33 & 82.83 & 67.00 \\
\bottomrule
\end{tabular}

\end{table}

\begin{table}[!htbp]
  \centering
  \small
  \setlength{\tabcolsep}{4pt}
  \caption{\textbf{Model-score similarity under direct Euclidean neighborhoods.}
  Similarity (\%) on the same $k=20$ neighbors as
  Table~\ref{tab:query-neighborhoods-direct-source}.}
  \label{tab:query-neighborhoods-direct-scores}
  \begin{tabular}{@{}lrrrrr@{}}
\toprule
Representation & EmbedLLM & NineBy30k & R2Bench & CARROT & Mean \\
\midrule
Random & 53.34 & 53.04 & 62.72 & 61.83 & 57.74 \\
E5-base & 60.56 & 58.75 & 71.10 & 70.10 & 65.13 \\
ModernBERT-base & 59.56 & 58.86 & 70.91 & 70.19 & 64.88 \\
Qwen2.5-0.5B & 61.39 & 59.70 & 71.35 & 70.02 & 65.62 \\
RegexRoute & 59.01 & 59.16 & 70.37 & 69.95 & 64.62 \\
\bottomrule
\end{tabular}

\end{table}

\clearpage
\section{Feature Recovery Across Encoder Sizes}
\label{app:feature-recovery}

\subsection{Setup and definitions}

We train linear ridge and nonlinear MLP probes to predict all 128 regex feature values from frozen query embeddings.
For each encoder and benchmark, we fit the probes on the training split, select hyperparameters by validation mean squared error, and evaluate on the test split.

\paragraph{Query representations.}
We use the seven Qwen2.5 sizes from 0.5B to 72B with the last-token embeddings described in Appendix~\ref{app:scaling}.
MiniLM-L6 and E5-base provide the smaller-encoder references in Figure~\ref{fig:feature-recovery-scaling}(c).
We also report ModernBERT-base in the tables below.
The Qwen2.5 models provide the comparison of scale within one model family.

\paragraph{Linear and MLP probes.}
Ridge regression uses the original embedding coordinates, with regularization selected from $\{1,10,\ldots,10^6\}$.
For MLP probe, we divide embeddings by their mean training-vector norm and fit a two-hidden-layer ReLU network with mean squared error loss.
Training uses AdamW, weight decay $10^{-4}$, batch size 512, and at most 80 epochs, with early stopping after eight epochs without validation improvement.
We search four configurations, specified as (hidden widths, dropout, learning rate): $((256,128),0.1,10^{-3})$, $((512,256),0.1,10^{-3})$, $((512,256),0.3,10^{-3})$, and $((1024,512),0.1,3\times10^{-4})$.
All runs use seed zero.

\paragraph{Recovery metric.}
For each feature $j$, we compare its predicted value $\widehat\phi_j(q)$ with its extracted value $\phi_j(q)$ across test queries:
\begin{equation}
R_j^2=1-\frac{\sum_{q\in\mathcal{D}_{\mathrm{route}}^{\mathrm{test}}}(\phi_j(q)-\widehat\phi_j(q))^2}
{\sum_{q\in\mathcal{D}_{\mathrm{route}}^{\mathrm{test}}}(\phi_j(q)-\overline\phi_{j,\mathrm{test}})^2},
\end{equation}
where $\overline\phi_{j,\mathrm{test}}$ is the feature's test mean.
A score of one denotes perfect recovery; zero matches predicting this mean for every query.

The main figure reports $R^2_{\mathrm{w}}=\sum_j w_jR_j^2/\sum_jw_j$ over these features.
Weights are impurity-based feature importances, plus $10^{-9}$, from an Extra-Trees regressor.
We also report the unweighted mean and median feature $R^2$.
Each summary is computed separately for each benchmark, then averaged equally across the four benchmarks.

\subsection{Linear and nonlinear recovery results}

Table~\ref{tab:feature-recovery} summarizes recovery across encoders, and Table~\ref{tab:feature-recovery-benchmarks} gives the per-benchmark results.
From Qwen2.5-0.5B to 72B, importance-weighted $R^2$ rises from 0.820 to 0.848 with an MLP, compared with 0.612 to 0.751 with ridge.
The small MLP gain also appears without importance weighting: mean feature $R^2$ rises from 0.734 to 0.761.
Thus, the surface features are already recoverable from the small encoder, while scaling improves linear recovery more substantially.

\begin{table}[!htbp]
\centering\small
\caption{Feature recovery across encoder sizes. Each entry averages four per-benchmark summaries. Wtd. uses training-derived Extra-Trees importances; mean and median treat features equally.}
\label{tab:feature-recovery}
\begin{tabular}{@{}lrrrrrr@{}}
\toprule
& \multicolumn{3}{c}{Ridge} & \multicolumn{3}{c}{MLP} \\
Encoder & Mean & Median & Wtd. & Mean & Median & Wtd. \\
\midrule
Qwen2.5-0.5B & 0.507 & 0.603 & 0.612 & 0.734 & 0.823 & 0.820 \\
Qwen2.5-1.5B & 0.542 & 0.642 & 0.652 & 0.751 & 0.843 & 0.840 \\
Qwen2.5-3B & 0.569 & 0.673 & 0.683 & 0.754 & 0.848 & 0.842 \\
Qwen2.5-7B & 0.594 & 0.701 & 0.713 & 0.754 & 0.854 & 0.841 \\
Qwen2.5-14B & 0.615 & 0.722 & 0.734 & 0.768 & 0.857 & 0.852 \\
Qwen2.5-32B & 0.581 & 0.718 & 0.727 & 0.748 & 0.841 & 0.833 \\
Qwen2.5-72B & 0.632 & 0.743 & 0.751 & 0.761 & 0.852 & 0.848 \\
MiniLM-22M & 0.345 & 0.393 & 0.424 & 0.689 & 0.770 & 0.751 \\
E5-base & 0.457 & 0.530 & 0.561 & 0.719 & 0.814 & 0.796 \\
ModernBERT-base & 0.563 & 0.662 & 0.673 & 0.696 & 0.790 & 0.774 \\
TF-IDF & 0.745 & 0.900 & 0.877 & --- & --- & --- \\
\bottomrule
\end{tabular}
\end{table}

\begin{table}[!htbp]
\centering\small
\caption{Importance-weighted feature recovery by benchmark. Probe configurations are selected separately on each benchmark\textquotesingle s validation split.}
\label{tab:feature-recovery-benchmarks}
\begin{tabular}{@{}lrrrr@{}}
\toprule
Encoder & EmbedLLM & NineBy30k & R2Bench & CARROT \\
\midrule
\multicolumn{5}{@{}l}{\textbf{Ridge (linear)}} \\
Qwen2.5-0.5B & 0.641 & 0.674 & 0.543 & 0.589 \\
Qwen2.5-1.5B & 0.667 & 0.721 & 0.589 & 0.632 \\
Qwen2.5-3B & 0.694 & 0.764 & 0.610 & 0.664 \\
Qwen2.5-7B & 0.740 & 0.775 & 0.644 & 0.692 \\
Qwen2.5-14B & 0.759 & 0.787 & 0.670 & 0.721 \\
Qwen2.5-32B & 0.773 & 0.776 & 0.661 & 0.699 \\
Qwen2.5-72B & 0.792 & 0.812 & 0.667 & 0.734 \\
MiniLM-22M & 0.570 & 0.422 & 0.323 & 0.380 \\
E5-base & 0.665 & 0.658 & 0.424 & 0.497 \\
ModernBERT-base & 0.792 & 0.723 & 0.561 & 0.618 \\
TF-IDF & 0.897 & 0.904 & 0.813 & 0.894 \\
\midrule
\multicolumn{5}{@{}l}{\textbf{MLP (nonlinear)}} \\
Qwen2.5-0.5B & 0.875 & 0.869 & 0.768 & 0.769 \\
Qwen2.5-1.5B & 0.882 & 0.868 & 0.775 & 0.835 \\
Qwen2.5-3B & 0.889 & 0.883 & 0.766 & 0.832 \\
Qwen2.5-7B & 0.887 & 0.878 & 0.772 & 0.829 \\
Qwen2.5-14B & 0.893 & 0.882 & 0.786 & 0.844 \\
Qwen2.5-32B & 0.896 & 0.881 & 0.747 & 0.807 \\
Qwen2.5-72B & 0.903 & 0.889 & 0.766 & 0.832 \\
MiniLM-22M & 0.881 & 0.672 & 0.673 & 0.776 \\
E5-base & 0.904 & 0.803 & 0.674 & 0.803 \\
ModernBERT-base & 0.902 & 0.818 & 0.584 & 0.792 \\
\bottomrule
\end{tabular}
\end{table}

\end{document}